# Supporting Autonomous Process Execution within a Multi-Perspective Constraint Frame via Numeric Planning

Paul Hermann Wittlinger[a], Giacomo Acitelli[b], Anti Alman[a], Fabrizio Maria Maggi[a], Andrea Marrella[b]

[a]*Free University of Bozen-Bolzano, Piazza Domenicani 3, 39100, Bolzano, Italy*
[b]*Sapienza Uniersity of Rome, via Ariosto 25, 00185, Rome, Italy*

**Abstract**

AI-Augmented Business Process Management Systems (ABPMS) enhance traditional BPMS by leveraging advanced AI techniques to define, execute, and monitor complex process structures. Within this landscape, *Framed Autonomy* denotes the capability of a system to autonomously advance the execution of a Business Process (BP) instance while strictly adhering to a predefined *frame*, i.e., a set of constraints that may span multiple perspectives. Existing research on framed autonomy has predominantly focused on control-flow constraints, either declarative or procedural, and typically relies on their transformation into automata-based representations. In this study, we extend this line of work by introducing a novel tool for *what-if* analysis that augments the process frame with multi-perspective constraints, including data-aware and temporal conditions. Given a partial process execution, the proposed approach exploits this enriched frame to recommend optimal continuations in compliance with the underlying process specifications. We additionally report an empirical evaluation demonstrating the scalability and effectiveness of the technique, thereby highlighting its potential for supporting autonomous and constraint-aware decision making in ABPMS.

*Keywords:* Business Process Management, Business Process Execution, Artificial Intelligence, Autonomous Agents, Automated Planning

---

*Email addresses:* pwittlinger@unibz.it (Paul Hermann Wittlinger), acitelli@diag.uniroma1.it (Giacomo Acitelli), anti.alman@unibz.it (Anti Alman), maggi@inf.unibz.it (Fabrizio Maria Maggi), marrella@diag.uniroma1.it (Andrea Marrella)

## 1. Introduction

The rising complexity of modern Information Systems has led to the emergence of AI-Augmented Business Process Management Systems (ABPMS) [1], extending traditional BPMS by incorporating Artificial Intelligence (AI) techniques, thereby enabling new forms of intelligent behavior. Specifically, ABPMS can *support* users by recommending suitable next actions, *reason* about potential root causes of observed behavior, and even *act autonomously* to progress ongoing Business Process (BP) instances. This autonomous decision-making capability, executed within well-defined operational boundaries, is referred to as *framed autonomy* [2].

A key element enabling framed autonomy is the notion of a *process frame*. A process frame is a holistic modeling construct that delineates the boundaries within which an ABPMS may operate. It integrates procedural or declarative process specifications with organizational rules, operational guidelines, and domain best practices. By combining multiple process perspectives, such as control flow, data, time, and resources, a process frame captures rich and potentially conflicting requirements that must be balanced during the execution of a process instance. Consequently, process frames support highly flexible system behaviors while ensuring adherence to domain-specific constraints.

Early work on framed autonomy has primarily focused on what-if analysis tools that compute admissible continuations of partial process executions. These partial executions may stem from real ongoing process instances or from hypothetical scenarios that a process analyst wishes to explore. In this context, the FrAIm tool [3] was introduced as a pioneering approach for framed autonomy. Given a prefix of an execution trace and a hybrid process frame composed of procedural models (Petri nets) and declarative constraints (Declare [4]), each associated with a violation cost, FrAIm employs classical planning techniques to generate suffixes that minimize the overall violation cost imposed by the frame.

Despite its contributions, FrAIm currently handles only control-flow perspectives and does not support data-aware or temporal constraints, which are essential for capturing the multi-perspective nature of real-world processes. To address this limitation, in this paper we extend the approach introduced through FrAIm by incorporating data and time conditions into the declarative component of the process frame, following the principles of Multi-Perspective Declare (MP-Declare [5]). This extension enables a richer and more expressive representation of process knowledge, thereby increasing the autonomy and decision-making capabilities of ABPMS in scenarios where compliance with multi-dimensional constraints is critical.

Our approach is grounded in *Numeric Planning*, a planning paradigm that natively supports numeric variables and arithmetic effects, thereby providing a natural mechanism for handling data attributes associated with process executions. Furthermore, existing extensions of Numeric Planning support *Temporal Planning*, enabling the representation of timestamps and the verification of metric temporal conditions, capabilities that we exploit to incorporate time-related MP-Declare constraints. A further advantage of numeric planners is their ability to optimize solutions with respect to multiple objectives. In our setting, we leverage this property to minimize both the overall violation cost of the process frame and the execution time of the process instance, thus balancing compliance with efficiency in the recommended continuations. Through an empirical evaluation on process data, we demonstrate the scalability and practical effectiveness of the proposed technique, showing that multi-perspective, data- and time-aware framed autonomy can be operationalized efficiently using Numeric Planning.

The remainder of this paper is structured as follows. Section 2 presents the preliminary knowledge required to understand the proposed approach. Section 3 introduces a running example that motivates the proposed solution and is used to illustrate the key concepts discussed throughout the paper. Section 4 describes our method for supporting autonomous process execution in a multi-perspective frame using Numeric Planning. Section 5 reports the experimental evaluation. Section 6 reviews the related work, and Section 7 concludes the paper and spells out directions for future research.

## 2. Preliminaries

In this section, we introduce the formal concepts needed to represent a multi-perspective process frame integrating procedural and declarative constraints. These notions will later serve as the foundation for the formalization of the multi-perspective framed autonomy problem. We also assume the availability of a previously observed (possibly empty) partial process execution.

### *2.1. Petri nets*

Petri nets [6] are a well-established formalism for modeling and analyzing discrete, state-based systems, and they have been extensively employed in business process modeling and verification [7]. A Petri net models the control-flow perspective through three core components: *transitions* (activities), *places* (states), and *arcs* (control-flow relations). Arcs connect places and transitions, and places may contain *tokens*. The distribution of tokens, called the *marking*, represents the current state of the process execution. Transitions update the marking when they *fire*: a transition consumes tokens from its input places and produces tokens in its output places, thereby modeling the execution of an activity. An example of a Petri net is provided in Figure 1.

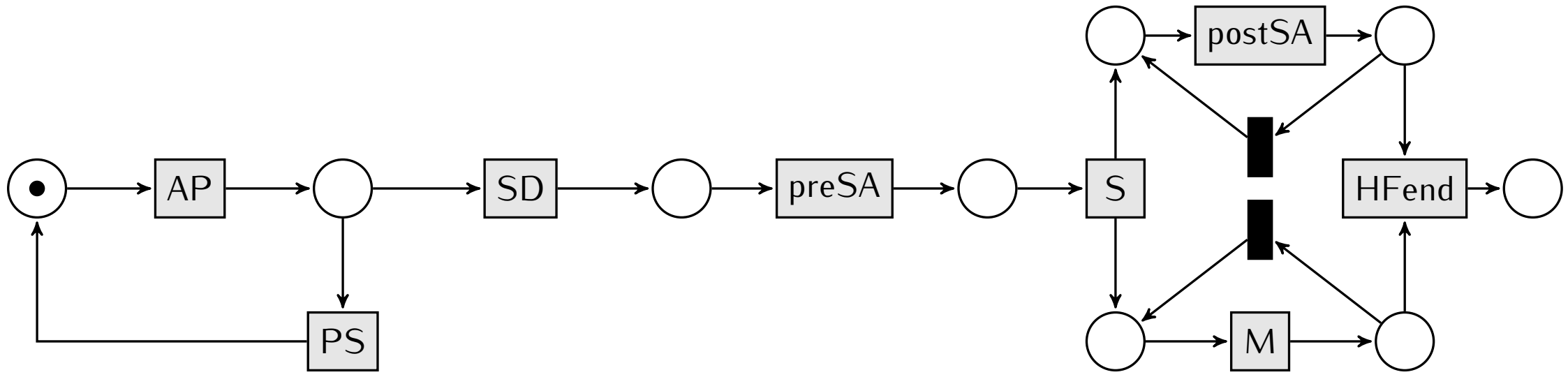

Figure 1: Petri net representing a CG procedure for hip fracture treatment.

**Definition 1** (Petri net). *A Petri net is a tuple* $\mathcal{N} = \langle P, T, F, M_0, M_f \rangle$*, where:*

- $P$ *is a finite set of places;*
- $T$ *is a finite set of (labeled) transitions;*
- $F \subseteq (P \times T) \cup (T \times P)$ *is the flow relation;*
- $M_0$ *is the initial marking;*
- $M_f$ *is the final marking.*

For a transition $t \in T$, the set of *input places*, denoted ${}^\bullet t$, is defined as $\{p \in P \mid (p, t) \in F\}$, and the set of *output places*, denoted $t^\bullet$, is $\{p \in P \mid (t, p) \in F\}$. A marking is a function $M : P \rightarrow \mathbb{N}$ assigning to each place the number of tokens it contains.

The operational semantics of a Petri net evolves as follows:

1. A transition $t$ is *enabled* in a marking $M$ if and only if each of its input places contains at least one token, i.e., for all $p \in {}^\bullet t$, $M(p) > 0$.
2. If $t$ is enabled, it may *fire*, yielding a new marking where one token is removed from each input place and one token is added to each output place.

In this paper, we focus on *1-bounded* Petri nets, i.e., nets in which every place contains at most one token at any time: $\forall p \in P,\ M(p) \leq 1$. The Petri net in our running example (Figure 1) belongs to this class. While we restrict data and temporal conditions to the declarative layer in this work, note that similar conditions can also be included in procedural models, such as in Data Petri Nets [8].

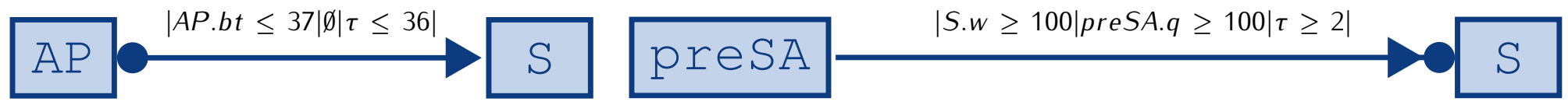

Figure 2: RESPONSE (left) and PRECEDENCE (right) modeling CG declarative constraints.

### 2.2. Multi-Perspective Declare

The DECLARE modeling language [4] is a declarative, constraint-based formalism grounded in Linear Temporal Logic over Finite Traces ($LTL_f$) [9]. Unlike procedural languages such as Petri nets, which explicitly prescribe execution paths, DECLARE specifies admissible behavior implicitly through constraints: any behavior that is not explicitly forbidden is permitted. A DECLARE model consists of a set of templates, each describing a reusable behavioral pattern that can be instantiated over an activity alphabet. The Multi-Perspective extension of DECLARE (MP-DECLARE) [5] enhances these templates by introducing data-aware and temporal conditions evaluated over activation and target events. This enables the specification of dependencies involving control-flow, event payload attributes, and timing constraints, thereby offering a more expressive and realistic formalism for process modeling.

An activation event $A$ occurring at time $\tau_A$ carries a payload $\varsigma(e_A)$, which stores the values of its attribute. A target event $B$ occurring at time $\tau_B$ carries a payload $\varsigma(e_B)$. An activation condition $\varphi_a$ is a predicate that must hold over the payload of the activation event for the constraint to be triggered, whereas a target condition $\varphi_t$ must hold over the payload of the target for the constraint to be satisfied. Throughout this paper, expressions such as $AP.bt$, $S.w$, and $preSA.q$ refer to specific attributes stored in the payload of the corresponding event: $AP.bt$ denotes the value of attribute $bt$ (body temperature) inside the payload of event AP, $S.w$ denotes the value of attribute $w$ (weight) inside the payload of event S, and $preSA.q$ denotes the value of attribute $q$ (anesthetic dosage) inside the payload of event preSA.

Temporal reasoning is defined using a subset of Metric First-Order Temporal Logic (MFOTL), first introduced in [10]. MFOTL extends classical temporal operators by annotating them with intervals that constrain the time window in which the embedded formula must hold. For instance, the future operator $\mathbf{F}_I$ requires that its argument eventually becomes true at some time point whose distance from the current one lies within the interval $I$. For example, when $I = [0, x]$, the formula $F_{[0,x]}\,\psi$ stipulates that $\psi$ must be satisfied at some point no later than $x$ time units from now. Analogous interpretations apply to the past and globally operators when equipped with metric intervals. These metric annotations provide precise control over temporal requirements such as deadlines, minimum waiting times, and bounded response conditions, which are essential for capturing the semantics of timed MP-Declare constraints used in this work. A complete account of MFOTL semantics is provided in [10].

In this paper, $I$ takes one of the forms $[0, \infty)$, $[x, \infty)$, or $[0, x]$, and is abstractly represented by a time condition $\varphi_\tau$ over a time variable $\tau$. This condition restricts the temporal distance (measured in hours)[1] between activation and target events. For example, the condition $\tau \leq 36$ requires the target to occur at a time distance of at most 36 hours from the activation, whereas $\tau \geq 2$ requires the target to occur at a time distance of least 2 hours from the activation. Table 1 reports the MFOTL semantics of all MP-Declare templates considered in this work, illustrating how activation conditions, target conditions, and temporal constraints jointly define the satisfaction of each constraint.

To illustrate how these elements interact, consider the constraints shown in Figure 2. We use the notation $|\varphi_a|\varphi_t|\varphi_\tau|$ to indicate activation, target, and temporal conditions ($\emptyset$ indicates no condition). The Response constraint on the left formalizes the recommendation that surgery (S) must occur within 36 hours after the assessment of the patient (AP), provided that the patient has regular body temperature. This is encoded by an activation condition $\varphi_a$ requiring $AP.bt \leq 37$, meaning that the constraint is triggered only when the payload of AP contains a body temperature not exceeding 37 degrees. The temporal condition $\varphi_\tau$ specifies the 36-hour window, while the target condition is trivially true.

The Precedence constraint on the right captures the dependency between surgery and the administration of anesthesia for patients with weight at least 100 kilograms. The activation condition $\varphi_a$ requires $S.w \geq 100$, meaning that the constraint is activated only when the payload of event S indicates a patient weight of at least 100 kg. The target condition $\varphi_t$ requires $preSA.q \geq 100$, ensuring that the dosage recorded in the payload of event preSA is at least 100 micrograms. The temporal condition $\varphi_\tau$ expresses that anesthesia must be administered at least two hours before surgery.

### *2.3. 1-clock Guarded Finite State Automata*

The approach to framed autonomy introduced in [11] employs deterministic finite-state automata (DFAs) to uniformly represent the procedural and declarative components of a process frame (i.e., Petri nets and Declare constraints, respectively). This translation provides a common semantic foundation for handling heterogeneous modeling formalisms. In the context of business processes, a DFA evaluates finite process executions, typically represented as sequences of events, each labeled with an activity, by processing the corresponding sequence of activity labels through a unique run of states, thereby accepting or rejecting the execution.

[1] In this paper we use hours as the time unit, but our solution supports any configurable time granularity.

Table 1: Semantics for MP-Declare constraints.

| Template | MFOTL Semantics |
|---|---|
| existence | $\mathbf{F}_I(A \wedge \varphi_a)$ |
| absence | $\neg\, \mathbf{F}_I(A \wedge \varphi_a)$ |
| responded existence | $\mathbf{G}((A \wedge \varphi_a) \rightarrow (\mathbf{O}_I(B \wedge \varphi_t) \vee \mathbf{F}_I(B \wedge \varphi_t)))$ |
| response | $\mathbf{G}((A \wedge \varphi_a) \rightarrow \mathbf{F}_I(B \wedge \varphi_t))$ |
| alternate response | $\mathbf{G}((A \wedge \varphi_a) \rightarrow \mathbf{X}(\neg(A \wedge \varphi_a)\mathbf{U}_I(B \wedge \varphi_t)))$ |
| chain response | $\mathbf{G}((A \wedge \varphi_a) \rightarrow \mathbf{X}_I(B \wedge \varphi_t))$ |
| precedence | $\mathbf{G}((A \wedge \varphi_a) \rightarrow \mathbf{O}_I(B \wedge \varphi_t))$ |
| alternate precedence | $\mathbf{G}((A \wedge \varphi_a) \rightarrow \mathbf{Y}(\neg(A \wedge \varphi_a)\mathbf{S}_I(B \wedge \varphi_t)))$ |
| chain precedence | $\mathbf{G}((A \wedge \varphi_a) \rightarrow \mathbf{Y}_I(B \wedge \varphi_t))$ |
| not responded existence | $\mathbf{G}((A \wedge \varphi_a) \rightarrow \neg(\mathbf{O}_I(B \wedge \varphi_t) \vee \mathbf{F}_I(B \wedge \varphi_t)))$ |
| not response | $\mathbf{G}((A \wedge \varphi_a) \rightarrow \neg\, \mathbf{F}_I(B \wedge \varphi_t))$ |
| not precedence | $\mathbf{G}((A \wedge \varphi_a) \rightarrow \neg\, \mathbf{O}_I(B \wedge \varphi_t))$ |
| not chain response | $\mathbf{G}((A \wedge \varphi_a) \rightarrow \neg\, \mathbf{X}_I(B \wedge \varphi_t))$ |
| not chain precedence | $\mathbf{G}((A \wedge \varphi_a) \rightarrow \neg\, \mathbf{Y}_I(B \wedge \varphi_t))$ |

In particular, any 1-bounded Petri net can be translated into its *reachability graph*, which is a DFA whose states correspond to the reachable markings of the net and whose edges encode the firing of transitions [12]. Likewise, every $\text{LTL}_f$ formula (and therefore any propositionalized MP-Declare constraint) can be compiled into a DFA that recognizes exactly the set of event sequences satisfying the formula [13]. This yields a uniform, automata-theoretic perspective in which Petri nets are mapped to reachability automata and MP-Declare constraints to temporal-logic automata, motivating the adoption of a single automaton model capable of capturing control-flow, data, and time.

To this end, we represent all components of a multi-perspective process frame $\mathcal{F}$ using *1-clock Guarded Finite State Automata* (1-GFAs). A 1-GFA extends a classical DFA by annotating its transitions with: (i) an activity label, (ii) a conditional data guard $\varphi$ evaluated over event payloads, and (iii) a temporal guard $\tau$ evaluated over a single clock. When all guards are trivial, namely, when each $\varphi$ evaluates to $\top$ and each $\tau$ imposes no restriction (e.g., $0 \leq \tau \leq \infty$), the 1-GFA behaves exactly like a standard DFA. This property allows us to encode both Petri nets and MP-Declare constraints within a unified formalism while retaining the expressiveness required to capture data- and time-dependent behavior.

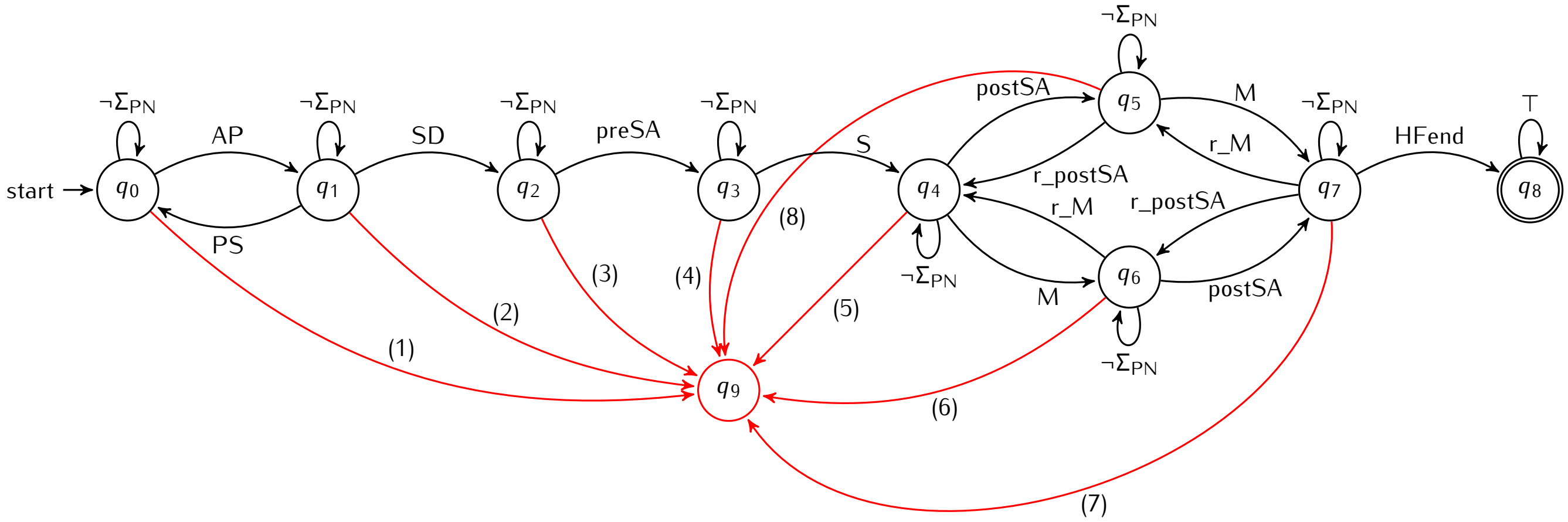

Figure 3: Procedural automaton of the CG represented as a Petri net in Figure 1. The labels r_M and r_postSA denote the silent transitions enabling the repetition of activities M and postSA. $\Sigma_{\mathrm{PN}}$ is the set of activities appearing in the Petri net. A negated set of labels indicates that all labels not belonging to that set are allowed. (1) = ¬AP; (2) = ¬{PS, SD}; (3) = ¬preSA; (4) = ¬S; (5) = ¬ { M, postSA }; (6) = ¬ { postSA, r_M }; (7) = ¬ { HFend, r_M, r_postSA }; (8) = ¬ { M, r_postSA }. Each guard constraint $\varphi \models \top$ and clock constraint $0 \leq \tau \leq \infty$.

**Definition 2** (1-Clock Guarded Finite State Automaton). *A 1-Clock Guarded Finite State Automaton (1-GFA) is a tuple $\mathcal{A} = \langle Q, x, \Sigma, \Delta, q^0, F \rangle$, where:*

- *$Q$ is a finite set of states;*
- *$x$ is a clock;*
- *$\Sigma$ is a finite alphabet of activity labels;*
- *$\Delta$ is a finite set of transitions;*
- *$q^0 \in Q$ is the initial state;*
- *$F \subseteq Q$ is a non-empty set of final states.*

Each transition $\delta \in \Delta$ has the form $\langle q, \mathtt{A}, \varphi, \tau, q', r \rangle$, where $q$ and $q'$ are the source and target states, $\mathtt{A} \in \Sigma$ is the activity label, $\varphi$ is a data guard over event payload attributes, $\tau$ is a temporal guard over the clock, and $r \in \{true, false\}$ indicates whether the clock is reset upon taking the transition. When an event labelled A occurs, the transition is enabled only if both $\varphi$ and $\tau$ hold; otherwise, the automaton cannot traverse $\delta$ and must remain in its current state or follow another enabled transition.

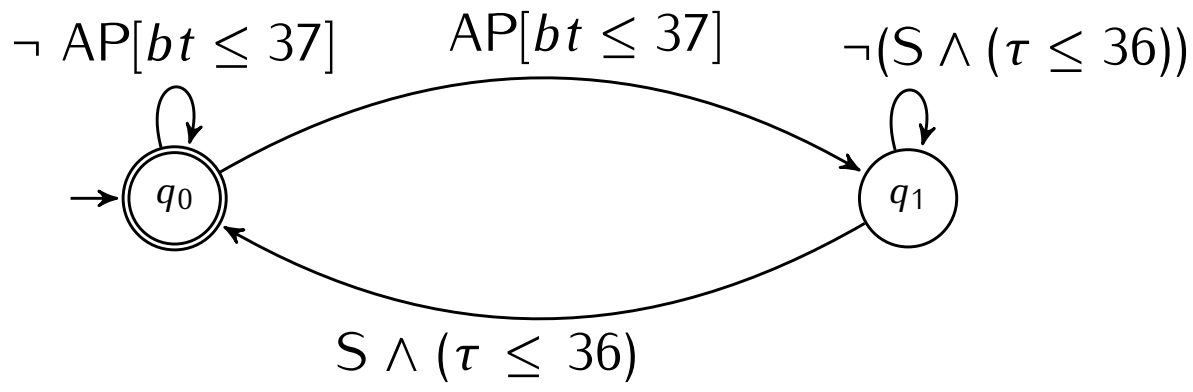

Figure 4: Declarative automaton of the Response constraint in Figure 2.

Returning to the running example, both the Petri net in Figure 1 and the MP-Declare constraints in Figure 2 can be translated into 1-GFAs, as illustrated in Figures 3 to 5. We refer to the 1-GFA obtained from the Petri net as the *procedural automaton*, denoted $\mathcal{A}_{\mathcal{P}}$, and to each 1-GFA derived from an MP-Declare constraint as a *declarative automaton*, denoted $\mathcal{A}_{\mathcal{D}}$. The 1-GFA encoding of the Petri net contains only trivial guards, as it captures control-flow ordering without data or temporal restrictions. Taken together, the procedural automaton and the set of declarative automata form a multi-perspective process frame $\mathcal{F}$, which integrates the procedural and declarative behavior of the system and will later serve as the basis for defining the temporal multi-perspective framed autonomy problem.

In addition to the components of the process frame $\mathcal{F}$, our what-if analysis tool may take as input a partial execution represented as a *prefix trace*. A prefix trace $t$ is a finite sequence of timed events describing a partially observed business process execution. Let $\Sigma_t$ be the set of activity labels observed in the prefix trace. Each event $e_i \in t$ is a tuple $\langle \mathtt{A}, p, ts \rangle$, where $\mathtt{A} \in \Sigma_t$ is the activity label, $p$ is the event payload, and $ts$ is the timestamp. We denote these projections by $\lambda(e_i)$, $\varsigma(e_i)$, and $\vartheta(e_i)$, respectively, so that $\lambda(e_i) = \mathtt{A}$, $\varsigma(e_i) = p$, and $\vartheta(e_i) = ts$.

Payloads are partial functions $p \in V^K$, where $K$ is a finite set of attributes and $V$ a finite domain of values. Missing values are indicated using a special symbol $\varepsilon \notin V$. Timestamps satisfy the strict ordering $\vartheta(e_1) < \ldots < \vartheta(e_n)$, ensuring the temporal consistency of the observed execution.

**Definition 3** ((Prefix) Trace)**.** *Let $\Sigma_t$ be a finite set of activity labels and $V^K$ a finite set of payloads. A (prefix) trace is a finite sequence $t = \langle e_1, \ldots, e_n \rangle$ such that $\lambda(e_i) \in \Sigma_t$, $\varsigma(e_i) \in V^K \cup \{\varepsilon\}$, and $\vartheta(e_1) < \cdots < \vartheta(e_n)$.*

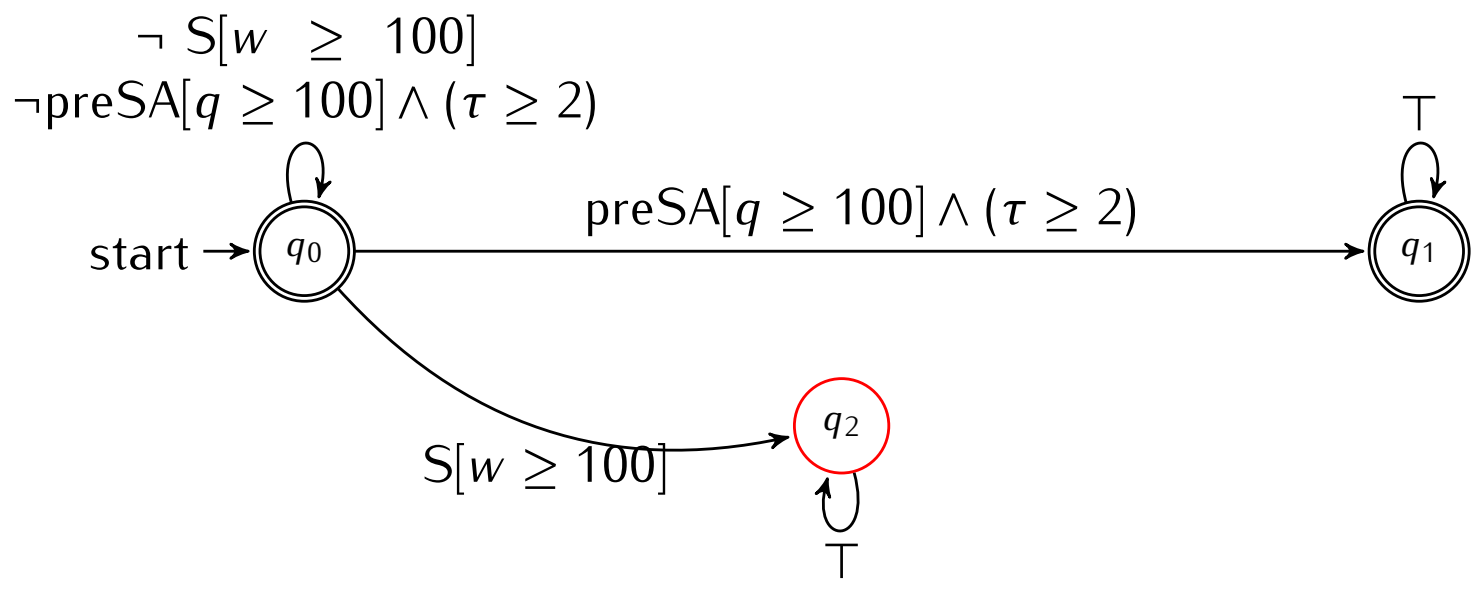

Figure 5: Declarative automaton of the PRECEDENCE constraint in Figure 2.

For automated reasoning, a prefix trace is also represented as a 1-GFA that encodes exactly the observed sequence.

**Definition 4** (Prefix Automaton)**.** *Given a prefix trace $t = \langle e_1, \ldots, e_n \rangle$ with labels in $\Sigma_t$, the corresponding prefix automaton is the 1-GFA $\mathcal{A}_t = \langle Q_t, x_t, \Sigma_t, \Delta_t, q_{0,t}, F_t \rangle$, where:*

- *$Q_t = \{q_0, \ldots, q_n\}$ is the sequence of states corresponding to the observed events;*
- *$x_t$ is the clock;*
- *$\Sigma_t$ is the input alphabet;*
- *$\Delta_t = \{\langle q_i, \lambda(e_{i+1}), \top, \top, q_{i+1}, false \rangle \mid i = 0, \ldots, n-1\}$ contains one transition per observed event, each with trivial guards;*
- *$q_t^0 = q_0$ is the initial state;*
- *$F_t = \{q_n\}$ is the final (accepting) state.*

An example prefix automaton $\mathcal{A}_t$ corresponding to the prefix trace $t = \langle (\texttt{AP}, \{bt = 37\}, 0), (\texttt{SD}, \emptyset, 4), (\texttt{Xray}, \emptyset, 16) \rangle$ is shown in Figure 6. The language accepted by $\mathcal{A}_t$, denoted $\mathcal{L}_{\mathcal{A}_t}$, consists of all event sequences beginning with this prefix.

**Definition 5** (Execution of a Prefix Trace on a Process Frame)**.** *Let $t = \langle e_1, \ldots, e_n \rangle$ be a prefix trace and $\mathcal{A}_t$ its prefix automaton. Let $\mathcal{F} = \{\mathcal{A}^1_{\mathcal{P}}, \cdots \mathcal{A}^h_{\mathcal{P}}, \mathcal{A}^1_{\mathcal{D}}, \cdots \mathcal{A}^k_{\mathcal{D}}\}$ be a process frame composed of h procedural automata and k declarative automata. An* execution *of $\rho$ over $\mathcal{F}$ is a sequence $q_0 \xrightarrow{e_1} q_1 \cdots \xrightarrow{e_n} q_n$ such that each transition is valid both in $\mathcal{A}_t$ and in every automaton in $\mathcal{F}$. We say that t* is accepted *by $\mathcal{F}$ (denoted $t \models \mathcal{F}$) if such an execution exists and the final state of $\mathcal{A}_t$ is accepting.*

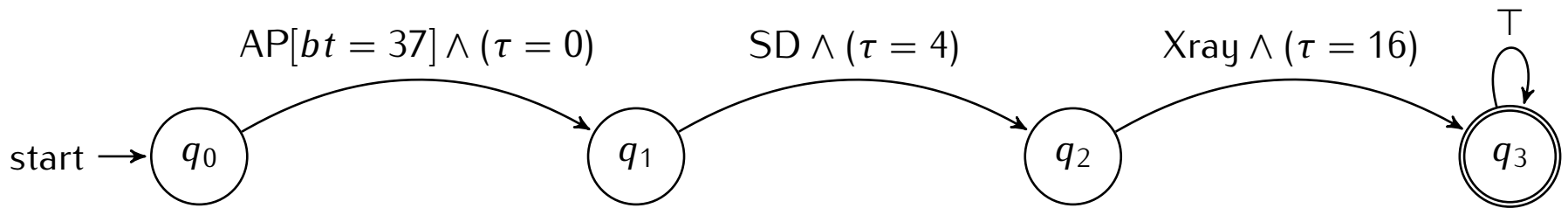

Figure 6: Prefix automaton for $t = \langle(\text{AP}, \{bt = 37\}, \tau = 0), (\text{SD}, \emptyset, \tau = 4), (\text{Xray}, \emptyset, \tau = 16)\rangle$.

### *2.4. Numeric Planning*

Automated Planning is a branch of Artificial Intelligence (AI) concerned with computing sequences of actions that transform a system from an initial state into one that satisfies a desired goal condition [14]. Given a description of the environment (i.e., the *planning domain*), a set of *action schemas* specifying possible state transitions, and a *planning problem* consisting of an initial state and a goal specification, a planning system searches the state space and returns an ordered sequence of actions (i.e., a plan) that achieves the goal when executed. The Planning Domain Definition Language (PDDL) [15] provides a standardized, declarative syntax for describing such domains and problems, enabling interoperability across different planning engines.

A key strength of Automated Planning is its generality: planners operate as domain-independent solvers, avoiding the need to design specialized ad-hoc algorithms for each new decision-making scenario. This has led to numerous applications in Business Process Management (BPM), including process adaptation, prediction, and automated decision support [16, 17, 18].

In this work, we rely on *numeric planning*, an extension of classical planning in which states include numeric fluents (variables ranging over $\mathbb{Q}$) that enable reasoning over quantities such as time, cost, and resource levels. When numeric planning is enriched with clock variables and temporal predicates, many planners become capable of supporting fragments of *temporal planning*, allowing the specification and verification of metric temporal constraints over action durations and ordering. These capabilities are essential in our setting, since the multi-perspective process frame must capture time-dependent behavioral requirements, such as those expressed by MP-Declare constraints (e.g., deadlines, minimum waiting times, or temporal separations between activities).

By leveraging temporal numeric planning, our approach is able to operationalize *framed autonomy*: given a partial process execution and a process frame that integrates procedural specifications with data-aware and time-aware declarative constraints, a planner synthesizes continuations that are both compliant with the frame and optimized with respect to user-defined criteria.

## 3. Running example

To illustrate the contribution of this work, we adopt the clinical guideline (CG) scenario introduced in [2], derived from recommendations issued by the British National Institute for Health and Care Excellence[2]. Figure 1 depicts a Petri net modeling a simplified treatment pathway for patients with a *hip fracture*. The process begins with an initial *assessment* (AP), during which the physician records the patients physical condition. If the assessment suggests that complications may arise during surgery, the physician may decide to *postpone* the surgery (PS), after which the patient must undergo a new assessment. Otherwise, a *surgery decision* (SD) is made and the intervention is scheduled. Once the surgery is planned, the procedure awaits the administration of pre-surgery anesthesia (preSA), which must be delivered before transporting the patient to the operating room, where the surgery (S) takes place. Following surgery, the patient receives post-surgical analgesia (postSA) and begins physiotherapy for mobilization (M). These activities may repeat until recovery, after which the patient is discharged (HFend).

Although the Petri net in Figure 1 captures the high-level sequencing of activities specified in the CG, it omits several clinically relevant dependencies between activities, particularly those involving time- and data-related conditions. For example, the guideline states that surgery should be performed "on the day of, or the day after, admission," i.e., within 36 hours [19]. Clinical decisions are also influenced by the patients physiological indicators. A patient with an elevated body temperature (bt), e.g., above 37°C, may require postponement of surgery due to possible comorbidities requiring immediate attention. Similarly, the effectiveness of pre-surgery anesthesia depends on both dosage and timing. For instance, if a dose of 1 $\mu$g/kg is required, surgeries for obese patients (e.g., with weight $w \geq 100$ kg) must be preceded by the administration of at least 100 $\mu$g, and this dosage must be delivered no later than two hours before surgery.

To incorporate these clinically meaningful constraints, we extend the procedural model with a set of MP-Declare constraints, shown in Figure 2. The Response constraint on the left formalizes the recommendation that surgery (S) must occur within 36 hours after the assessment, provided that the patient is otherwise healthy ($bt \leq 37$). This requirement is encoded through an activation condition on the attribute *bt* in the payload of AP (denoted as $AP.bt$). The Precedence constraint on the right captures the dependency between surgery performed on obese patients ($S.w \geq 100$) and the administration of anesthesia (preSA), together with their temporal distance. Specifi-cally, it requires that the anesthesia dosage satisfies $preSA.q \geq 100$ (expressed as a target condition) and that anesthesia is administered sufficiently in advance (at least two hours before surgery).

[2] http://www.nice.org.uk/

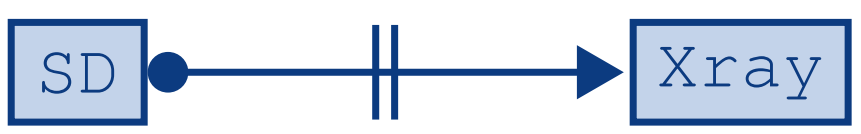

Figure 7: Not Response constraint modeling the Basic Medical Knowledge.

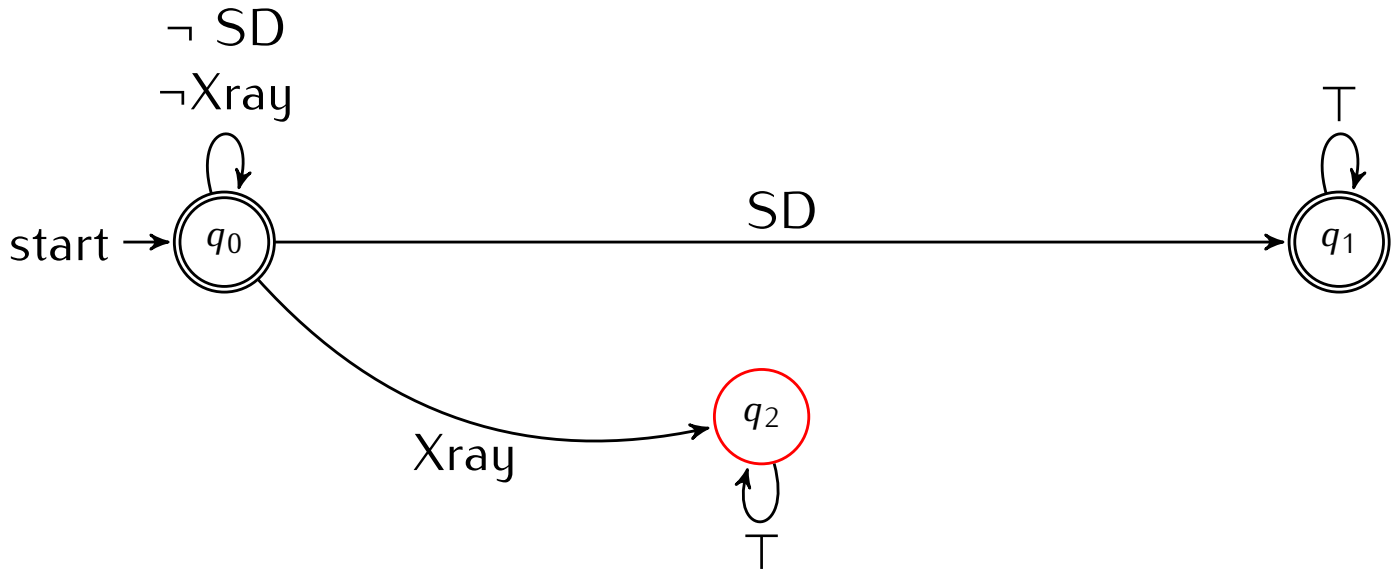

Figure 8: Declarative automaton of the Not Response constraint in Figure 7.

Clinical guidelines are typically defined under the assumption of an idealized execution context [20]. This implies, among other things, unlimited availability of resources, the absence of comorbidities other than those explicitly addressed by the guideline, and complete clinical knowledge on the part of the healthcare provider regarding both the medical conditions involved and the patient to whom the guideline applies. While such assumptions are acceptable when CGs are used for documentation, education, or model-driven analysis, they become overly restrictive when guidelines are enacted at runtime or employed for decision support. In real clinical settings, physicians must manage the full complexity of individualized care: patients present heterogeneous medical conditions, personal constraints and preferences, comorbidities, and unexpected complications. These factors are governed by what is referred to as *background medical knowledge* [20], which introduces additional decision points and branches conditioned on diagnostic evidence.

Consider, for example, the initial assessment (AP) in the running scenario. If the patient shows symptoms suggestive of a chest infection (e.g., fever, cough), the physician may prescribe a chest X-ray (Xray). Should the X-ray reveal a severe infection (e.g., persistent or severe cough), the guideline indicates that the infection be treated with priority, postponing hip surgery until the condition stabilizes. In this situation, the patient must receive amoxicillin therapy (AT), and the hip-fracture pathway is temporarily suspended. If instead the infection is mild, amoxicillin therapy may be omitted, and the surgical procedure may proceed once the patient is re-evaluated and considered stable. Crucially, the surgery decision (SD) must not be taken before all relevant diagnostic information is available; therefore, the activity Xray must not occur *after* SD. This requirement can be captured by a Declare constraint of type Not Response, as shown in Figures 7 and 8, together with its corresponding 1-GFA representation.

Taken together, the procedural specification (the Petri net) and the declarative specifications (the MP-Declare constraints) form a *hybrid business process representation* (HBPR) [21]. The HBPR captures both procedural dependencies and declarative constraints that incorporate data-aware and time-aware conditions, enabling the modeling of complex and context-sensitive clinical decision pathways. In this paper, the HBPR constitutes the foundation for the autonomous reasoning mechanism we propose: given a partial execution and the multi-perspective process frame, the system determines which next activities are both feasible and compliant, thereby supporting safe and informed clinical decision making under realistic, non-ideal execution conditions.

## 4. Supporting Autonomous Process Executions

In this section, we introduce the formal definition of multi-perspective framed autonomy, show how to encode its components into 1-clock guarded finite-state automata, and finally describe how the resulting problem can be reduced to a numeric temporal planning problem.

### 4.1. Multi-Perspective Framed Autonomy

Consider a prefix trace $t = \langle e_1, \dots, e_n \rangle$ over the alphabet $\Sigma_t$, and let $\mathcal{A}_t$ be its corresponding prefix automaton. Let $\mathcal{F}$ be a multi-perspective process frame composed of $h$ Petri nets, represented by the procedural automata $\mathcal{A}^1_{\mathcal{P}}, \dots, \mathcal{A}^h_{\mathcal{P}}$, and $k$ MP-Declare constraints, represented by the declarative automata $\mathcal{A}^1_{\mathcal{D}}, \dots, \mathcal{A}^k_{\mathcal{D}}$. We denote by $\mathcal{A}_{\mathcal{F}} = \{\mathcal{A}^1_{\mathcal{P}}, \dots, \mathcal{A}^h_{\mathcal{P}}, \mathcal{A}^1_{\mathcal{D}}, \dots, \mathcal{A}^k_{\mathcal{D}}\}$ the set of all automata in the frame.

**Definition 6** (Multi-Perspective Process Frame Automata Set)**.** *Given a multi-perspective process frame $\mathcal{F}$, the associated* frame automata set *is $\mathcal{A}_{\mathcal{F}} = \{\mathcal{A}^1_{\mathcal{P}}, \dots, \mathcal{A}^h_{\mathcal{P}}, \mathcal{A}^1_{\mathcal{D}}, \dots, \mathcal{A}^k_{\mathcal{D}}\}$, with alphabet $\Sigma_{\mathcal{F}} = \bigcup_{i=1}^{h} \Sigma_{\mathcal{A}^i_{\mathcal{P}}} \cup \bigcup_{i=1}^{k} \Sigma_{\mathcal{A}^i_{\mathcal{D}}}$.*

As in classical framed autonomy, in the multi-perspective setting we aim to generate a suffix trace $\hat{t}$ extending $t$ such that the concatenation $t \cdot \hat{t}$ yields a complete execution that minimally violates the constraints in $\mathcal{F}$. To achieve this, we first define what it means for a suffix of $t$ to be valid within $\mathcal{F}$.

**Definition 7** (Valid Suffix within a Process Frame). *Given a prefix trace $t = \langle e_1, \dots, e_n \rangle$ with prefix automaton $\mathcal{A}_t$ and a multi-perspective process frame $\mathcal{F}$ with $\mathcal{A}_\mathcal{F} = \{\mathcal{A}^1_\mathcal{P}, \dots, \mathcal{A}^h_\mathcal{P}, \mathcal{A}^1_\mathcal{D}, \dots, \mathcal{A}^k_\mathcal{D}\}$, a* valid suffix *within $\mathcal{F}$ is a trace $\hat{t} = \langle e_{n+1}, \dots, e_{n+m} \rangle$ such that there exists an execution $q_0 \xrightarrow{e_1} q_1 \cdots \xrightarrow{e_n} q_n \xrightarrow{e_{n+1}} q_{n+1} \cdots \xrightarrow{e_{n+m}} q_{n+m}$ for which, for every $i = 0, \dots, n+m-1$ and every automaton $\mathcal{A} \in \mathcal{A}_\mathcal{F} \cup \{\mathcal{A}_t\}$, the transition $\langle q_i, e_{i+1}, q_{i+1} \rangle \in \Delta_\mathcal{A}$ exists and each such automaton ends in an accepting state.*

The above definition states that $\hat{t}$ is a valid suffix of $t$ within $\mathcal{F}$ exactly when the concatenation $t \cdot \hat{t}$ is accepted by $\mathcal{F}$, written $t \cdot \hat{t} \models \mathcal{F}$.

To ensure full expressive power during suffix generation, the alphabet of each automaton $\mathcal{A}_j \in \mathcal{A}_\mathcal{F}$ is extended with all reset activities reset($\mathcal{A}_i$) for every $\mathcal{A}_i \in \mathcal{A}_\mathcal{F}$, yielding the augmented alphabet $\Sigma^+_{\mathcal{A}_j}$. Each activity reset($\mathcal{A}_i$) represents an operation that, when executed, forces the corresponding automaton $\mathcal{A}_i$, and its associated clock, back to its initial state, while leaving the remaining automata unchanged. This mechanism is essential for completeness. In many situations, constructing a *valid* suffix that satisfies all automata in $\mathcal{F}$ may be impossible because of conflicting procedural, data-aware, or temporal constraints. Reset activities offer a principled way to resolve such conflicts: by selectively resetting an automaton, it is still possible to continue constructing a suffix that minimally violates the process frame. For this reason, reset activities are incorporated into every augmented alphabet and become available whenever it is determined that violating $\mathcal{A}_j$ is unavoidable in order to obtain a completion of minimum cost.

We define:

- $\Sigma_\mathcal{F} = \bigcup_{j=1}^{h} \Sigma_{\mathcal{A}^j_\mathcal{P}} \ \cup \ \bigcup_{j=1}^{k} \Sigma_{\mathcal{A}^j_\mathcal{D}}$;
- $\Sigma = \Sigma_\mathcal{F} \cup \Sigma_t$;
- $\Sigma^+ = \bigcup_{j=1}^{h} \Sigma^+_{\mathcal{A}^j_\mathcal{P}} \ \cup \ \bigcup_{j=1}^{k} \Sigma^+_{\mathcal{A}^j_\mathcal{D}} \cup \Sigma_t$.

Activities in $\Sigma^+$ are called *repair activities*. Among these, the activities contained in $\Sigma$ always incur zero cost, while the reset actions reset($\mathcal{A}_j$) are assigned a configurable positive cost, as they explicitly represent violations of the automata in $\mathcal{A}_\mathcal{F}$. A sequence over $\Sigma^+$ is called a *repair trace*. Intuitively, a repair trace specifies the adjustments required to transform the prefix $t$ into a complete execution that minimally violates $\mathcal{F}$. This may occur either by identifying a valid suffix that is directly accepted by $\mathcal{F}$, or, when this is not possible, by generating a suffix $\hat{t}$ induced by a repair trace of minimum cost. The suffix is referred to as the *optimal continuation* of the prefix.

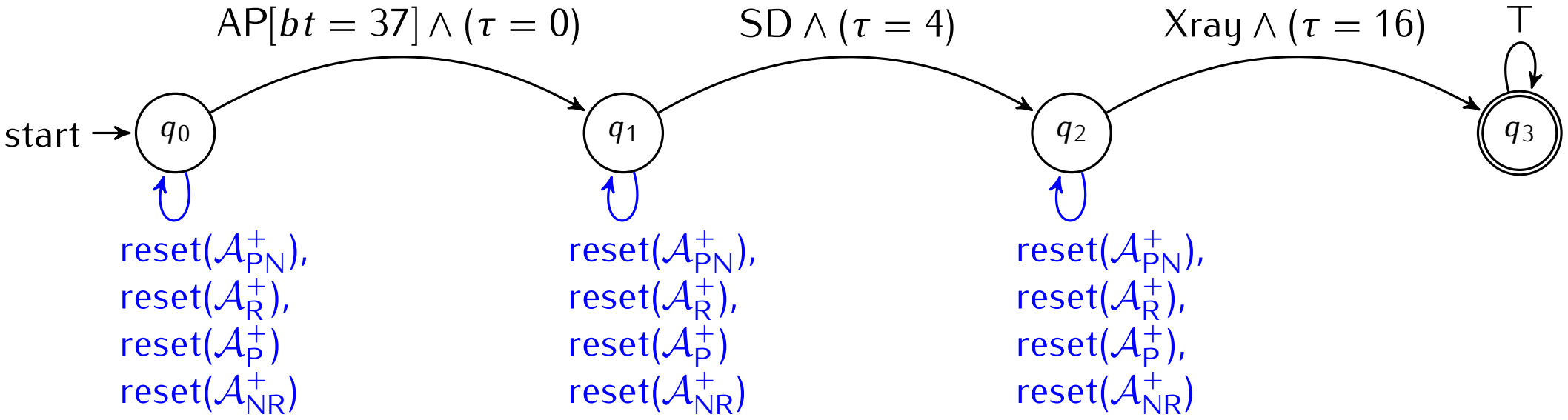

Figure 9: Augmented prefix automaton $\mathcal{A}^+_t$ for for $t = \langle(\mathrm{AP}, \{bt = 37\}, \tau = 0), (\mathrm{SD}, \emptyset, \tau = 4), (\mathrm{Xray}, \emptyset, \tau = 16)\rangle$.

To extend a prefix trace with an optimal continuation, we need to generate an applicable repair trace, and then derive from it the optimal continuation.

**Definition 8** (Applicable repair trace). *A repair trace $t^+$ is* applicable *to the prefix $t$ if:*

- *any number of activities* reset($\mathcal{A}_j$) *are inserted between events of $t$;*
- *or new events are appended to $t$.*

**Definition 9** (Induced trace). *Given an applicable repair trace $t^+$, the induced trace $t^*$ is obtained by removing all occurrences of* reset($\mathcal{A}_j$) *from $t^+$. We call $t^*$ the* optimal continuation *of $t$ if $t^+$ is a minimum-cost repair trace.*

*4.2. Automata-theoretic approach*

In order to account for all activities in $\Sigma^+$, the automata in $\mathcal{A}_\mathcal{F}$ and the prefix automaton $\mathcal{A}_t$ must be augmented with transitions corresponding to the repair activities reset($\mathcal{A}_j$). The following definition introduces the augmented prefix automaton.

**Definition 10** (Augmented Prefix Automaton). *Let $\mathcal{A}_t = \langle Q_t, x_t, \Sigma_t, \Delta_t, q^0_t, F_t\rangle$ be a prefix automaton. Its* augmented prefix automaton *is defined as $\mathcal{A}^+_t = \langle Q_t, x_t, \Sigma^+_t, \Delta^+_t, q^0_t, F_t\rangle$, where $\Delta^+_t$ extends $\Delta_t$ by adding:*

*(i) a self-loop $\langle q, \mathrm{reset}(\mathcal{A}_j), \top, (0 \leq \tau \leq \infty), q\rangle$ for every $q \in Q_t$ and every $\mathcal{A}_j \in \mathcal{A}_\mathcal{F}$;*

*(ii) a transition $\langle q_n, \Sigma^+, \top, (0 \leq \tau \leq \infty), q_n, false\rangle$, enabling terminal synchronization.*

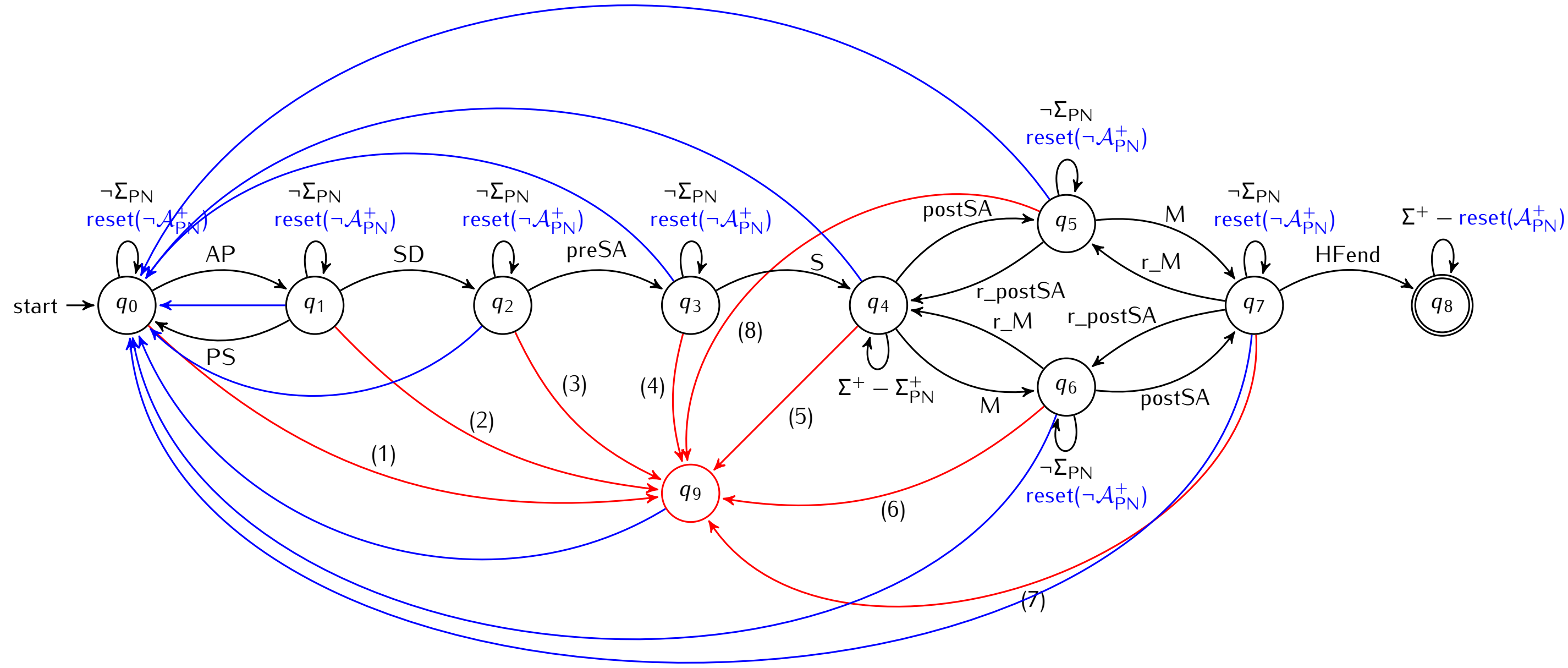

Figure 10: Augmented procedural automaton of the CG represented as a Petri net in Figure 1, where: (1) = ¬AP; (2) = ¬{PS, SD}; (3) = ¬ { preSA }; (4) = ¬ { S }; (5) = ¬ { M, postSA }; (6) = ¬ { postSA, r_M }; (7) = ¬ { HFend, r_M, r_postSA }; (8) = ¬ { M, r_postSA }. The blue arcs are the newly introduced reset arcs all corresponding to the repair activity reset($\mathcal{A}^+_{\text{PN}}$).

Intuitively, reset operations do not alter the state evolution of the prefix automaton and therefore appear as self-loops. Consequently, the language of the original prefix automaton is contained in that of its augmented version, i.e., $\mathcal{L}_{\mathcal{A}_t} \subseteq \mathcal{L}_{\mathcal{A}_t^+}$. Figure 9 illustrates the augmented prefix automaton derived from the running example. Here, $\mathcal{A}^+_{PN}$, $\mathcal{A}^+_R$, $\mathcal{A}^+_P$, and $\mathcal{A}^+_{NR}$ denote the augmented automata corresponding to the Petri net and to the Response, Precedence, and Not Response constraints, respectively, of our running example, within the augmented frame defined as follows.

**Definition 11** (Augmented Frame). *Let $\mathcal{F}$ be a multi-perspective process frame, and let $\mathcal{A}_j = \langle Q_{\mathcal{A}_j}, x_{\mathcal{A}_j}, \Sigma_{\mathcal{A}_j}, \Delta_{\mathcal{A}_j}, q_0^{\mathcal{A}_j}, F_{\mathcal{A}_j} \rangle$ be any automaton in $\mathcal{A}_{\mathcal{F}}$. Its* augmented automaton *is the structure*

$$\mathcal{A}_j^+ = \langle Q_{\mathcal{A}_j}, x_{\mathcal{A}_j}, \Sigma^+_{\mathcal{A}_j}, \Delta^+_{\mathcal{A}_j}, q^0_{\mathcal{A}_j}, F_{\mathcal{A}_j}, cost(\mathcal{A}_j), cost(\tau) \rangle,$$

*where:*

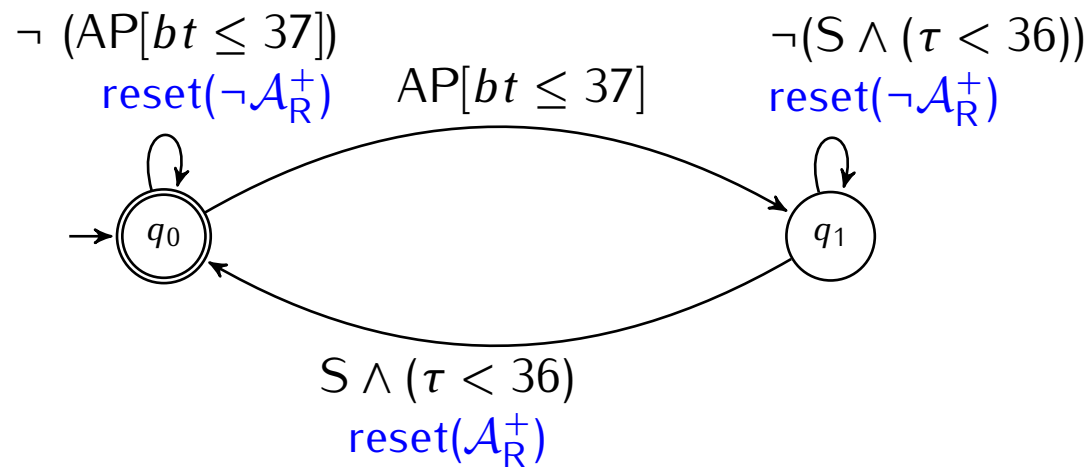

Figure 11: Augmented declarative automaton of the Response constraint in Figure 2.

- $\Delta^+_{\mathcal{A}_j}$ *extends* $\Delta_{\mathcal{A}_j}$ *with:*
  - *(i) a reset transition* $\langle q, \mathsf{reset}(\mathcal{A}_j), \top, (0 \leq \tau \leq \infty), q^0_{\mathcal{A}_j}, \mathit{true}\rangle$ *for all* $q \in Q_{\mathcal{A}_j} \setminus (F_{\mathcal{A}_j} \cup \{q^0_{\mathcal{A}_j}\})$*;*
  - *(ii) a self-loop* $\langle q, \mathsf{reset}(\mathcal{A}_i), \top, (0 \leq \tau \leq \infty), q, \mathit{false}\rangle$ *for all* $q \in Q_{\mathcal{A}_j}$ *and for all* $\mathcal{A}_i \in \mathcal{A}_{\mathcal{F}}$ *with* $i \neq j$*.*
- $\mathit{cost}(\mathcal{A}_j)$ *assigns a non-negative cost to executing* $\mathsf{reset}(\mathcal{A}_j)$*;*
- $\mathit{cost}(\tau)$ *represents the cost of advancing the clock* $x_{\mathcal{A}_j}$ *by one time unit.*

*The augmented frame* $\mathcal{F}^+$ *is the frame whose automata are given by:* $\mathcal{A}_{\mathcal{F}^+} = \{\bigcup_{j=1}^{h} \mathcal{A}^j_{\mathcal{P}} \cup \bigcup_{j=1}^{k} \mathcal{A}^j_{\mathcal{D}}\}$.

Figures 10 to 13 depict the augmented automata corresponding to the procedural model and the declarative constraints derived from the running example. Here the notation $\mathsf{reset}(\neg\mathcal{A}_j)$ indicates the reset of any automaton in $\mathcal{A}_{\mathcal{F}}$ except $\mathcal{A}_j$.

We are now ready to state the core problem addressed in this work.

**Definition 12** (Multi-perspective Framed Autonomy)**.** *Let* $t$ *be a prefix trace, and let* $\mathcal{F} = \{\mathcal{A}^1_{\mathcal{P}}, \ldots, \mathcal{A}^h_{\mathcal{P}}, \mathcal{A}^1_{\mathcal{D}}, \ldots, \mathcal{A}^k_{\mathcal{D}}\}$ *be a multi-perspective process frame. Multi-perspective framed autonomy is the optimization problem of finding a repair trace* $t^+$ *over* $\Sigma^+$ *such that:*

- $t^+$ *is applicable to* $t$*;*
- *the induced trace* $t^*$ *satisfies* $t^* \models \mathcal{F}^+$*;*
- *after replaying* $t^+$*, every automaton in* $\mathcal{A}_{\mathcal{F}^+}$ *is in an accepting state;*
- *the cost of* $t^+$ *is minimal.*

In the next section, we provide an automata-theoretic formalization that enables solving this optimization problem through numeric temporal planning.

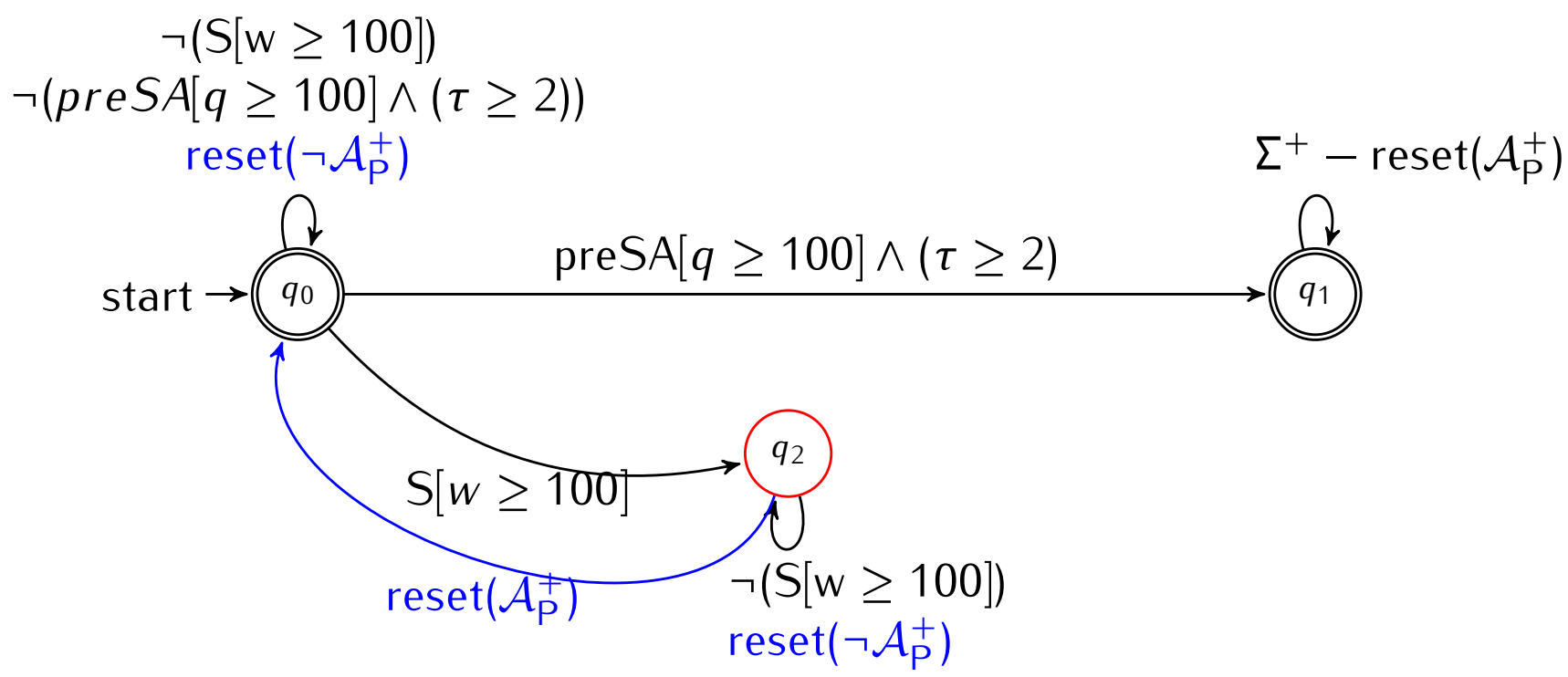

Figure 12: Augmented declarative automaton of the PRECEDENCE constraint in Figure 2.

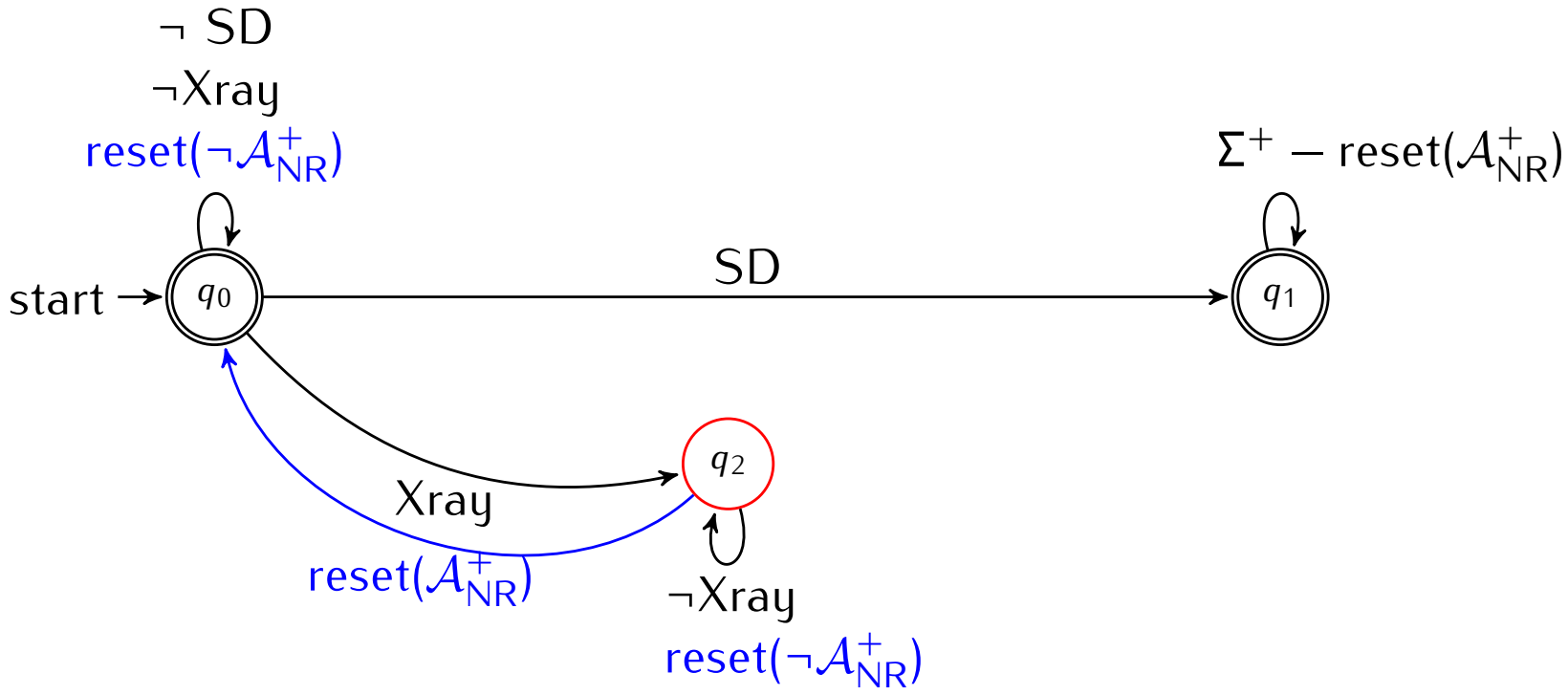

Figure 13: Augmented declarative automaton of the NOT RESPONSE constraint in Figure 7.

### *4.3. Numeric Planning for Framed Autonomy*

Based on the previously defined approach, we now show how to encode the multi-perspective framed autonomy problem as a numeric planning problem. Planning actions are constructed from the activities in the augmented alphabet $\Sigma^+$. The execution of any planning action synchronously advances both the augmented trace automaton $\mathcal{A}_t^+$ and each augmented automaton $\mathcal{A}_j^+ \in \mathcal{A}_{\mathcal{F}^+}$.

Planning actions are executable only when permitted by the current state of the automata: an action $a$ may be executed if, and only if, the current states of $\mathcal{A}_j^+ \in \mathcal{A}_{\mathcal{F}^+}$ and $\mathcal{A}_t^+$ have an outgoing transition labeled with $a$. The task is therefore to determine a deterministic plan that takes these 1-GFAs from their initial states to final states at minimal cost. Actions corresponding to activities in $\Sigma^+$ are assigned costs according to a user-defined cost function $\mathcal{C}$. A plan generated in this way represents a complete execution trace that solves the framed autonomy problem, i.e., a sequence of actions that extends the prefix trace with a suffix satisfying $\mathcal{F}$ at minimum cost. The details of this reduction are provided below.

A *numeric planning problem* over a set of Boolean propositions $\Phi$ and continuous variables $\Psi$ is a tuple $\mathcal{D} = \langle F, X, \Omega, s_0, cost, G \rangle$, where:

- $F \subseteq 2^{\Phi}$ is a finite set of Boolean variables;
- $X$ is a finite set of continuous variables over $\mathbb{Q}^{\Psi}$;
- $\Omega$ is a finite set of *domain actions*;
- $cost : \Omega \mapsto \mathbb{N}_0$ is a *cost function* assigning a non-negative cost to each action;
- $G$ is the goal condition.

A *plan* $\pi$ for $\mathcal{D}$ is a finite sequence $\pi = a_1 \cdots a_n \in \Omega^*$ of actions, which is said to be *executable* in $\mathcal{D}$ from a state $s_0$ if there exists a sequence of states $\sigma = s_0 \cdots s_n$ such that, for all $i = 0, \ldots, n-1$, $s_{i+1} = \tau(s_i, a_{i+1})$. If such a sequence exists, it is called the *domain trace induced by* $\pi$. Since $\mathcal{D}$ is deterministic, every plan uniquely determines a domain trace. The *cost* of a plan is defined as $cost(\pi) \doteq \sum_{i=1,\ldots,n} cost(a_i)$.

A *cost-optimal planning problem* is a tuple $\mathcal{P} = \langle \mathcal{D}, s_0, G \rangle$, where:

- $\mathcal{D}$ is a planning domain with action costs;
- $s_0$ is the *initial state*;
- $G$ is the *goal*, a propositional formula over $\Phi \times \Psi$.

A plan $\pi$ is a *solution* if the last state $s_n$ of its induced trace satisfies $s_n \models G$. A solution $\pi$ is *optimal* if, for every other solution $\pi'$, it holds that $cost(\pi) \leq cost(\pi')$.

We encode the multi-perspective framed autonomy problem as follows. Let $t$ be a prefix trace and $\mathcal{F}$ a multi-perspective process frame with automata $\mathcal{A}_{\mathcal{F}} = \{\mathcal{A}^1_{\mathcal{P}}, \ldots, \mathcal{A}^h_{\mathcal{P}}, \mathcal{A}^1_{\mathcal{D}},$
$\ldots, \mathcal{A}^k_{\mathcal{D}}\}$. The current states of $\mathcal{A}_j \in \mathcal{A}_{\mathcal{F}}$ and of $\mathcal{A}_t$ are represented by Boolean propositions $F \subseteq 2^{\Phi}$ and fluents over $X \subset \mathbb{Q}^{\Psi}$. Note that the states of the augmented automata coincide with those of their non-augmented versions (Definitions 10 and 11). We also assume that $Q_t$ and $Q_{\mathcal{A}_j}$ are disjoint for all $\mathcal{A}_j \in \mathcal{A}_{\mathcal{F}}$.

We define the planning domain $\mathcal{D} = \langle S, \Omega, cost, \tau \rangle$, where:

1. $S \subseteq \{ \{ q_t \} \cup \{ \bigcup_{j=1,\ldots,h+k} q_{\mathcal{A}_j} \} \mid q_t \in Q_t, q_{\mathcal{A}_j} \in Q_{\mathcal{A}_j} \}$

2. $\Omega = \Sigma^+$, i.e., every activity in $\Sigma^+$ is a planning action. An event $e_i$ models a synchronous step executed across all framing automata. All components of $e_i$ (i.e., $\lambda(e_i)$, $\varsigma(e_i)$, and $\vartheta(e_i)$) contribute to the transition. Reset actions $\text{reset}(\mathcal{A}_j) \in \Sigma^+_{\mathcal{A}_j}$ model resets of the framing automata;
3. for all $A \in \Sigma$, $cost(A) = 0$, whereas $cost(\text{reset}(\mathcal{A}_j))$ is defined by the user-specified cost function $\mathcal{C}$;
4. the transition function is defined as follows: for $\omega \in \Omega$, $q_t, q'_t \in Q_t$, and $q_{\mathcal{A}_j}, q'_{\mathcal{A}_j} \in Q_{\mathcal{A}_j}$,

$$\delta(\{q_t\} \cup \{\bigcup_{j=1,\dots,h+k} q_{\mathcal{A}_j}\}, \omega) = \{q'_t\} \cup \{\bigcup_{j=1,\dots,h+k} q'_{\mathcal{A}_j}\}$$

   if and only if:
   (a) $\langle q_t, \omega, q'_t \rangle \in \Delta^+_t$;
   (b) $\langle q_{\mathcal{A}_j}, \omega, q'_{\mathcal{A}_j} \rangle \in \Delta^+_{\mathcal{A}_j}$ for each $\mathcal{A}_j \in \mathcal{A}_{\mathcal{F}}$.

We refer to $\mathcal{D}$ as the *framed autonomy planning domain* associated with $t$ and $\mathcal{F}$. It models the synchronous product of $\mathcal{A}^+_t$ and $\mathcal{A}^+_j$ for all $\mathcal{A}_j \in \mathcal{A}_{\mathcal{F}}$.

Since the domain is deterministic, each state of $\mathcal{D}$ contains exactly one state from $Q_t$ and one state from each $Q_{\mathcal{A}_j}$, ensuring a unique successor for every executable action $\omega$. Conditions (a) and (b) above encode executability: an action $\omega$ may be executed only if it is enabled in the current state of $\mathcal{A}^+t$ and of every $\mathcal{A}^+_j$.

Now we define the cost-optimal planning problem $\mathcal{P} = \langle \mathcal{D}, s_0, G \rangle$, where:

- $s_0 = \{q^0_t, q^0_{\mathcal{A}^+_1}, q^0_{\mathcal{A}^+_2}, \dots\}$ is the initial state;
- $G = \left(\bigwedge_{q_{\mathcal{A}_j} \in F_{\mathcal{A}_j}} q_{\mathcal{A}_j}\right) \wedge q_n$ is the goal condition, where $q_n$ is the unique final state of $\mathcal{A}^+_t$. For each $\mathcal{A}_j$, the set $F_{\mathcal{A}_j}$ corresponds to the final states of its non-augmented version.

We call $\mathcal{P}$ the *multi-perspective framed autonomy planning problem* associated with $t$ and $\mathcal{F}$. A solution is a minimal-cost plan whose induced domain trace ends in a state satisfying $G$. As noted above, such plans correspond exactly to repair traces.

## 5. Evaluation

In this section, we report on a set of experiments designed to evaluate the approach introduced in Section 4. To this end, we developed a Java-based tool that transforms a multi-perspective process frame $\mathcal{F}$, together with a (potentially empty) prefix trace $t$, into a numeric planning problem.[3] The tool takes as input: *(i)* an MP-Declare specification (in *.decl* format), *(ii)* a Petri net (in *.pnml* format), and *(iii)* a (potentially empty) prefix trace (in *.xes* format). In addition, it accepts a user-defined cost model specifying, for each constraint, its violation cost (in *.txt* format). If no cost model is provided, every reset operation is assigned a default cost of 1.

### *5.1. Experimental Setup*

The scalability of the proposed approach depends on several factors, including the complexity of the procedural models, the number of declarative constraints, and the length of the prefix trace. Consequently, the experimental setup is necessarily articulated. All tests rely on three inputs: a Petri net, an MP-Declare specification, and a prefix trace. Below, we describe how each input is generated, followed by an overview of the complete test design.

We employ four Petri nets in total. We begin with a relatively simple base model, generated using PLG2 [22]. This model, named '0And', contains 13 activities, no loops, and three XOR-structures, one of which is nested (i.e., located within a branch of another XOR). We then derive three progressively more parallel variants of this model by converting one XOR structure at a time into an AND structure. These variants are denoted '1And', '2And', and '3And'. This allows us to assess how the approach scales with increasing parallelism, and thus with an enlarged state space, of the procedural model. The resulting models are reported in Appendix A.

Similarly, we employ four MP-Declare specifications. These specifications are derived from a hand-crafted ordered set of seven MP-Declare constraints,[4] each containing both data and time conditions. Any prefix of this ordered set yields a well-formed, nontrivial constraint model: specifically, we use the first $n_c$ constraints, with $n_c \in \{1, 3, 5, 7\}$. The constraints and their ordering are provided in Appendix A.

Combining the four Petri nets with the four MP-Declare specifications yields 16 possible process frames. For each frame, we generate three *satisfying* and three *violating* prefixes of length $n_t \in \{1, 3, 4\}$. Each prefix includes the timestamps and data payloads that are relevant for the declared constraints. Satisfying prefixes are obtained by taking the first $n_t$ activities of a valid execution of the corresponding Petri net, whereas violating prefixes are obtained by taking the last $n_t$ activities of such an execution. We also include an empty prefix for each process frame.[5] For violating prefixes, the induced trace must include at least $n$ repair actions of type *reset* during the replay of $t$, allowing us to assess the scalability of the approach under conflicting situations. For satisfying prefixes, the prefix length affects only the size of the suffix that must be generated.

---

[3] The tool is publicly available at: `https://github.com/pwittlinger/numeric-PDDL_generator/tree/data-framed-autonomy`

[4] Automatically generating these constraints is not feasible here, as they must be mutually consistent and meaningful across all Petri nets.

[5] A prefix can violate a process frame only if it contains at least one event.

| | | Grounding Time | | | | | | | | | | | | Search Time | | | | | | | | | | | |
|---|---|---|---|---|---|---|---|---|---|---|---|---|---|---|---|---|---|---|---|---|---|---|---|---|---|
| | | 1 Constraint | | | 3 Constraints | | | 5 Constraints | | | 7 Constraints | | | 1 Constraint | | | 3 Constraints | | | 5 Constraints | | | 7 Constraints | | |
| Model | $n_t$ | Data | Timed | Both | Data | Timed | Both | Data | Timed | Both | Data | Timed | Both | Data | Timed | Both | Data | Timed | Both | Data | Timed | Both | Data | Timed | Both |
| 0And | 0 | 2.49 | 2.35 | 2.30 | 4.76 | 4.50 | 4.84 | 7.52 | 7.53 | 8.04 | 13.54 | 12.51 | 13.43 | 0.04 | 0.04 | 0.04 | 0.04 | 0.05 | 0.04 | 0.04 | 0.08 | 0.04 | 0.08 | 0.14 | 0.07 |
| | 1 | 2.57 | 2.20 | 2.28 | 4.60 | 4.69 | 4.53 | 8.05 | 7.99 | 7.49 | 13.05 | 12.46 | 12.08 | 0.04 | 0.04 | 0.04 | 0.04 | 0.05 | 0.04 | 0.04 | 0.08 | 0.04 | 0.05 | 0.09 | 0.05 |
| | 3 | 2.35 | 2.23 | 2.21 | 4.96 | 5.01 | 5.02 | 8.37 | 8.40 | 8.48 | 12.99 | 12.76 | 12.91 | 0.10 | 0.10 | 0.10 | 0.10 | 0.09 | 0.10 | 0.13 | 0.21 | 0.12 | 0.16 | 0.35 | 0.16 |
| | 4 | 2.32 | 2.12 | 2.26 | 5.05 | 5.09 | 5.04 | 8.79 | 8.84 | 8.73 | 13.56 | 13.28 | 13.15 | 0.12 | 0.11 | 0.11 | 0.11 | 0.11 | 0.11 | 0.14 | 0.24 | 0.15 | 0.17 | 0.33 | 0.17 |
| 1And | 0 | 2.44 | 2.33 | 2.31 | 5.18 | 5.08 | 5.16 | 8.99 | 8.99 | 8.82 | 12.95 | 13.05 | 13.10 | 0.04 | 0.04 | 0.04 | 0.03 | 0.06 | 0.04 | 0.04 | 0.09 | 0.04 | 0.06 | 0.20 | 0.05 |
| | 1 | 2.47 | 2.44 | 2.45 | 5.34 | 5.28 | 5.30 | 9.56 | 9.72 | 9.47 | 14.24 | 14.06 | 13.97 | 0.04 | 0.04 | 0.04 | 0.04 | 0.06 | 0.03 | 0.04 | 0.09 | 0.04 | 0.06 | 0.13 | 0.08 |
| | 3 | 2.72 | 2.60 | 2.52 | 5.63 | 5.52 | 5.74 | 10.52 | 10.78 | 10.86 | 14.60 | 14.40 | 14.40 | 0.11 | 0.10 | 0.10 | 0.10 | 0.10 | 0.10 | 0.09 | 0.17 | 0.09 | 0.13 | 0.44 | 0.15 |
| | 4 | 2.76 | 2.59 | 2.65 | 5.79 | 5.72 | 5.86 | 10.97 | 11.25 | 11.28 | 14.85 | 14.91 | 14.89 | 0.11 | 0.11 | 0.11 | 0.11 | 0.12 | 0.11 | 0.11 | 0.20 | 0.11 | 0.16 | 0.40 | 0.15 |
| 2And | 0 | 3.77 | 3.58 | 3.68 | 9.42 | 9.32 | 9.38 | 13.90 | 14.07 | 13.57 | 20.67 | 20.07 | 19.51 | 0.05 | 0.05 | 0.05 | 0.04 | 0.07 | 0.04 | 0.06 | 0.13 | 0.06 | 0.07 | 0.19 | 0.06 |
| | 1 | 4.04 | 3.77 | 3.77 | 9.71 | 9.61 | 9.89 | 13.80 | 13.75 | 13.72 | 20.05 | 19.84 | 20.18 | 0.06 | 0.05 | 0.05 | 0.04 | 0.07 | 0.04 | 0.06 | 0.13 | 0.06 | 0.08 | 0.31 | 0.09 |
| | 3 | 4.09 | 3.94 | 3.99 | 10.37 | 10.33 | 10.43 | 14.81 | 15.14 | 14.98 | 22.04 | 22.04 | 21.56 | 0.09 | 0.09 | 0.09 | 0.09 | 0.11 | 0.09 | 0.11 | 0.25 | 0.11 | 0.15 | 0.42 | 0.15 |
| | 4 | 4.45 | 4.18 | 4.19 | 10.85 | 10.71 | 10.71 | 15.44 | 15.63 | 15.85 | 23.48 | 22.23 | 22.69 | 0.10 | 0.09 | 0.10 | 0.11 | 0.13 | 0.10 | 0.11 | 0.20 | 0.12 | 0.16 | 0.45 | 0.16 |
| 3And | 0 | 5.76 | 5.57 | 5.56 | 12.64 | 12.52 | 12.55 | 19.58 | 19.89 | 19.39 | 26.97 | 26.05 | 27.14 | 0.04 | 0.05 | 0.04 | 0.05 | 0.08 | 0.05 | 0.07 | 0.14 | 0.07 | 0.10 | 0.29 | 0.09 |
| | 1 | 5.84 | 5.73 | 5.69 | 13.12 | 13.01 | 13.27 | 19.73 | 20.18 | 20.33 | 28.23 | 27.99 | 27.95 | 0.05 | 0.05 | 0.05 | 0.05 | 0.11 | 0.05 | 0.06 | 0.12 | 0.06 | 0.09 | 0.36 | 0.10 |
| | 3 | 6.05 | 5.94 | 5.95 | 14.01 | 13.71 | 14.08 | 21.57 | 21.53 | 22.16 | 30.72 | 30.05 | 31.73 | 0.08 | 0.08 | 0.07 | 0.10 | 0.20 | 0.11 | 0.11 | 0.28 | 0.10 | 0.15 | 1.12 | 0.15 |
| | 4 | 6.86 | 6.87 | 6.77 | 14.16 | 14.20 | 14.15 | 22.35 | 22.89 | 22.58 | 32.09 | 33.52 | 32.47 | 0.09 | 0.10 | 0.09 | 0.10 | 0.21 | 0.10 | 0.11 | 0.49 | 0.11 | 0.18 | 0.97 | 0.20 |

Table 2: Experiments using the satisfying prefixes (including the empty prefix).

Each MP-Declare specification is further split into three variants: *(i)* containing only data conditions; *(ii)* containing only time conditions; and *(iii)* containing both. These variants are referred to as "data", "timed", and "both", respectively. No additional prefixes are required for these variants, since the Java implementation disregards any payload attributes (including timestamps) that are not referenced in the active constraints. However, the number of MP-Declare specifications increases from four to twelve.

In total, this results in 336 distinct configurations (4 Petri nets × 12 MP-Declare specifications × 7 prefixes). Each configuration is executed five times, and the average grounding and search times are reported in Table 2 and Table 3. All tests assign a reset cost of 1000 to each constraint and a waiting cost of 10 per time unit. Experiments were executed sequentially on a machine with 16 GB of RAM and an AMD Ryzen 4800U CPU, with a memory limit of 10 GB per test. The Expressive Numeric Heuristic Search Planner (ENHSP) [23], configured with the 'blind' heuristic (option *opt-blind*), was used in all experiments.

### 5.2. Results

The results reported in Table 2 and Table 3 are summarized in Figs. 14 and 15. In what follows, we rely primarily on the figures, as they more clearly convey the overall trends. The primary axis (left) reports the *search time* in seconds, while the secondary axis (right, opaque) reports the *grounding time*. The total runtime is the sum of these two components.

Each figure summarizes a specific experimental configuration. Both search and grounding time increase as the number of constraints grows from the first row (1 constraint) to the last row (7 constraints). Frames containing only temporal guards pose the greatest challenge for search. This is due to the less restrictive ACTIVATION conditions of the temporal constraints, which prevents them from being vacuously satisfied and consequently expands the search space.

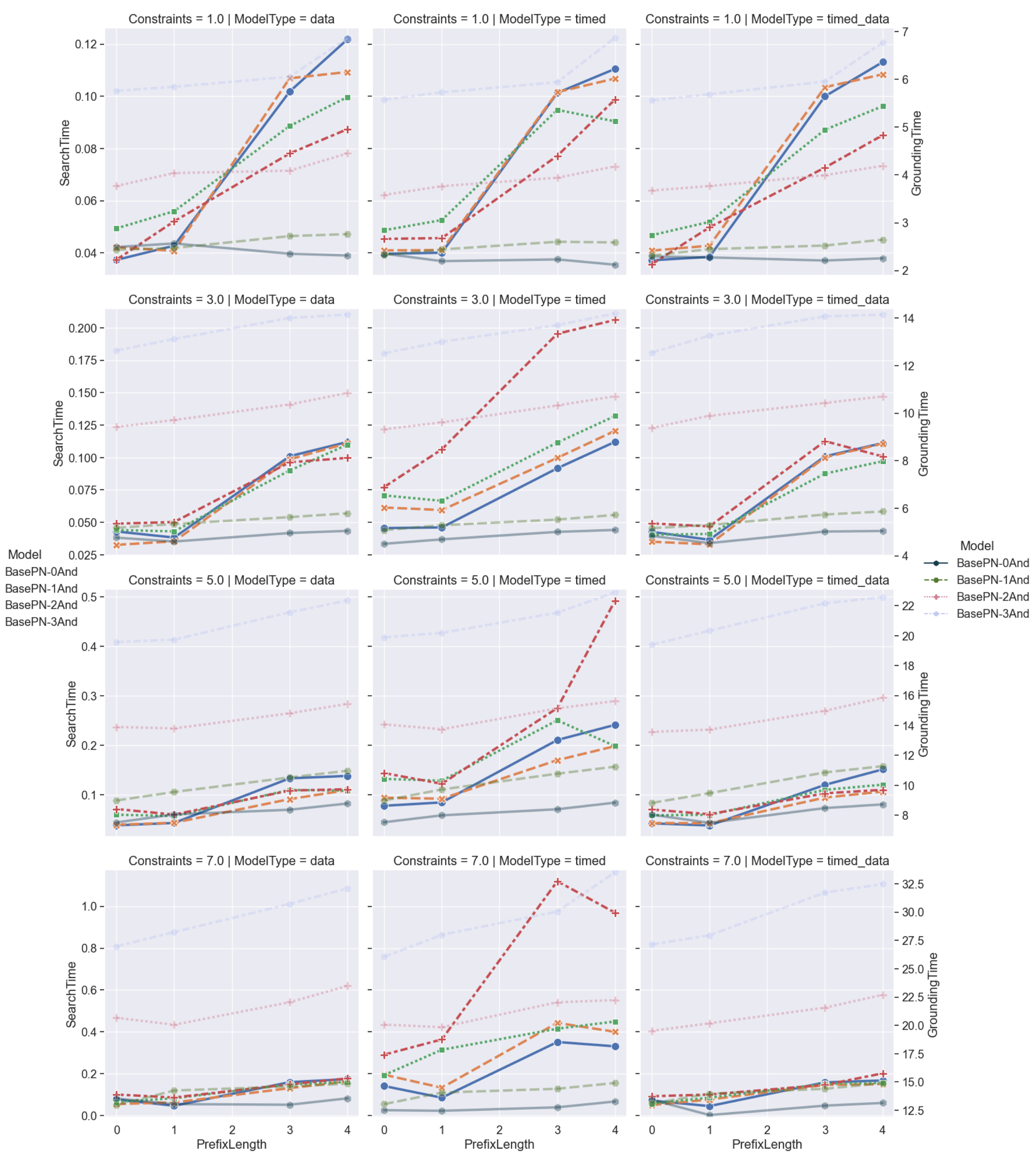

Figure 14: Average search time in seconds for a satisfying prefix.

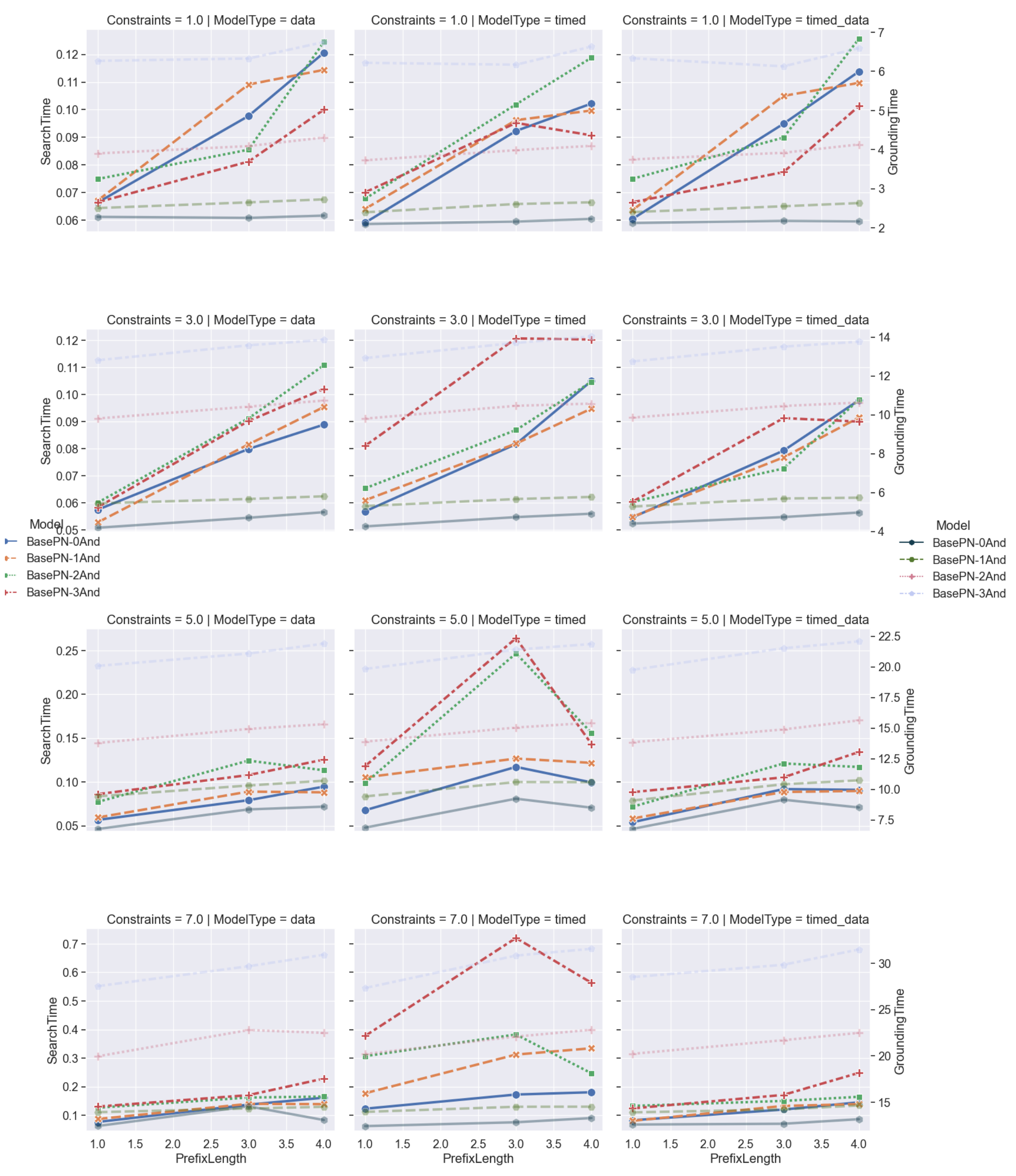

Figure 15: Average search time in seconds for a violating prefix.

| | | **Grounding Time** | | | | | | | | | | | | **Search Time** | | | | | | | | | | | |
|---|---|---|---|---|---|---|---|---|---|---|---|---|---|---|---|---|---|---|---|---|---|---|---|---|---|
| | | 1 Constraint | | | 3 Constraints | | | 5 Constraints | | | 7 Constraints | | | 1 Constraint | | | 3 Constraints | | | 5 Constraints | | | 7 Constraints | | |
| Model | $n_t$ | Data | Timed | Both | Data | Timed | Both | Data | Timed | Both | Data | Timed | Both | Data | Timed | Both | Data | Timed | Both | Data | Timed | Both | Data | Timed | Both |
| 0And | 1 | 2.28 | 2.10 | 2.12 | 4.19 | 4.26 | 4.40 | 6.77 | 6.88 | 6.78 | 12.44 | 12.43 | 12.60 | 0.07 | 0.06 | 0.06 | 0.06 | 0.06 | 0.05 | 0.06 | 0.07 | 0.05 | 0.08 | 0.12 | 0.08 |
| | 3 | 2.26 | 2.16 | 2.18 | 4.70 | 4.73 | 4.74 | 8.36 | 9.24 | 9.15 | 14.56 | 12.84 | 12.68 | 0.10 | 0.09 | 0.09 | 0.08 | 0.08 | 0.08 | 0.08 | 0.12 | 0.09 | 0.14 | 0.17 | 0.12 |
| | 4 | 2.31 | 2.23 | 2.17 | 4.99 | 4.91 | 4.97 | 8.59 | 8.50 | 8.52 | 13.08 | 13.30 | 13.17 | 0.12 | 0.10 | 0.11 | 0.09 | 0.10 | 0.10 | 0.09 | 0.10 | 0.09 | 0.16 | 0.18 | 0.15 |
| 1And | 1 | 2.51 | 2.40 | 2.40 | 5.44 | 5.30 | 5.28 | 9.43 | 9.43 | 9.09 | 13.93 | 13.96 | 13.91 | 0.07 | 0.06 | 0.06 | 0.05 | 0.06 | 0.05 | 0.06 | 0.11 | 0.06 | 0.09 | 0.18 | 0.08 |
| | 3 | 2.65 | 2.61 | 2.55 | 5.67 | 5.66 | 5.69 | 10.32 | 10.60 | 10.42 | 14.33 | 14.51 | 14.23 | 0.11 | 0.10 | 0.10 | 0.08 | 0.08 | 0.08 | 0.09 | 0.13 | 0.09 | 0.14 | 0.31 | 0.13 |
| | 4 | 2.73 | 2.65 | 2.63 | 5.80 | 5.77 | 5.73 | 10.70 | 10.59 | 10.74 | 14.51 | 14.52 | 14.66 | 0.11 | 0.10 | 0.11 | 0.10 | 0.09 | 0.09 | 0.09 | 0.12 | 0.09 | 0.14 | 0.33 | 0.14 |
| 2And | 1 | 3.90 | 3.73 | 3.75 | 9.81 | 9.80 | 9.85 | 13.79 | 13.90 | 13.85 | 19.96 | 20.20 | 20.23 | 0.07 | 0.07 | 0.07 | 0.06 | 0.07 | 0.06 | 0.08 | 0.10 | 0.07 | 0.13 | 0.31 | 0.13 |
| | 3 | 4.09 | 3.99 | 3.92 | 10.42 | 10.46 | 10.45 | 14.93 | 15.05 | 14.89 | 22.79 | 22.08 | 21.70 | 0.09 | 0.10 | 0.09 | 0.09 | 0.09 | 0.07 | 0.12 | 0.25 | 0.12 | 0.16 | 0.38 | 0.15 |
| | 4 | 4.31 | 4.10 | 4.13 | 10.73 | 10.57 | 10.62 | 15.33 | 15.44 | 15.64 | 22.49 | 22.83 | 22.49 | 0.12 | 0.12 | 0.13 | 0.11 | 0.10 | 0.10 | 0.11 | 0.16 | 0.12 | 0.17 | 0.25 | 0.16 |
| 3And | 1 | 6.27 | 6.22 | 6.34 | 12.81 | 12.92 | 12.75 | 20.08 | 19.83 | 19.76 | 27.53 | 27.33 | 28.51 | 0.07 | 0.07 | 0.07 | 0.06 | 0.08 | 0.06 | 0.09 | 0.12 | 0.09 | 0.13 | 0.38 | 0.12 |
| | 3 | 6.33 | 6.17 | 6.13 | 13.57 | 13.70 | 13.51 | 21.09 | 21.41 | 21.52 | 29.68 | 30.83 | 29.84 | 0.08 | 0.10 | 0.08 | 0.09 | 0.12 | 0.09 | 0.11 | 0.26 | 0.11 | 0.17 | 0.72 | 0.17 |
| | 4 | 6.75 | 6.63 | 6.59 | 13.86 | 14.03 | 13.76 | 21.90 | 21.87 | 22.09 | 30.92 | 31.56 | 31.49 | 0.10 | 0.09 | 0.10 | 0.10 | 0.12 | 0.09 | 0.13 | 0.14 | 0.13 | 0.23 | 0.56 | 0.25 |

Table 3: Experiments using violating prefixes.

Grounding time, by contrast, is largely unaffected by the type of constraints (i.e., data, timed, or both) and by the nature of the prefix. This is expected, as grounding depends mainly on the structural properties of the model. Across all experiments, search time remains negligible relative to grounding time.

Procedural parallelism has a more substantial effect on grounding time than on search time. Grounding consistently increases with the degree of parallelism and shows a modest increase with the prefix length.

Finally, note that many MP-DECLARE constraints may be vacuously satisfied, especially when enriched with data or temporal conditions. For example, a classical RESPONSE constraint on activities $A$ and $B$ is activated on every occurrence of $A$. If the same constraint includes the guard $x < 10$, it is activated only when $A$ occurs with a payload value below this threshold. This impacts the kind of plans returned by the planner. Since waiting incurs a cost, albeit lower than resetting, it may be preferable to avoid triggering a constraint's activation condition, thereby avoiding waiting cost altogether.

## 6. Related Literature

Within the BPM domain, several research branches have developed techniques to identify and uncover behavioral patterns in process models. One such line of work is *Trace Alignment* [24], which provides algorithms for reconciling process models with (potentially violating) business process executions. Trace alignment offers detailed feedback on how a given execution deviates from its reference model and, beyond quantifying the degree of deviation, proposes *repair strategies* to remedy the violation.

Alignment techniques have been successfully applied in several domains, including healthcare [25, 26, 27], the public sector [28, 29], and other application areas [30, 31, 32, 33, 34].

A variety of trace alignment methods exist [35, 36], including approaches grounded in Automated Planning. Conformance checking via planning [37, 38, 17, 39] typically relies on automata transformations to reduce the alignment problem to a classical planning task. In [40], numeric planning is further used to support the alignment of temporal constraints with event logs. Most existing work focuses on aligning either procedural models (e.g., Petri nets) [37, 41] or declarative specifications (e.g., MP-Declare constraints) [17, 39]. A unified approach for aligning event logs with respect to both paradigms was proposed by van Dongen et al. [42].

The combination of *procedural* and *declarative* formalisms to represent process behavior is referred to as Hybrid Business Process Representations (HBPR) [43]. This emerging field aims to formalize the interaction between heterogeneous, multi-perspective process models. While the general objective of trace alignment over HBPRs is related to that of *framed autonomy*, an important distinction remains: in framed autonomy, the observed prefix $t$ cannot be modified and must be replayed exactly as recorded, whereas trace alignment permits altering the observed behavior.

Predictive Process Monitoring (PPM) [44] complements framed autonomy in a related yet orthogonal manner. As in framed autonomy, PPM combines a process model with a partial execution prefix, but its objective is to *predict* the most likely continuation of the process based on historical event data rather than to generate a continuation that is guaranteed to comply with the process constraints. PPM therefore produces *probabilistic* suffixes, which implies that *(i)* predicted continuations may deviate from the reference model when such deviations are statistically more likely in past executions, and *(ii)* the next recommended actions are nondeterministic, reflecting the underlying probability distribution rather than a compliance requirement. In contrast, the planning-based framed autonomy introduced in this paper is inherently *prescriptive*: it synthesizes continuations that must satisfy the multi-perspective process frame (or minimally violate it through explicitly costed repair actions), thus ensuring constraint-driven correctness rather than likelihood-driven prediction. As a result, PPM and planning-based framed autonomy serve complementary purposes, forecasting likely future behavior versus computing compliant, optimal continuations, each supporting different forms of intelligent assistance within AI-augmented Business Process Management Systems.

## 7. Conclusion and Future Work

In this paper, we presented an automata-based formalization of multi-perspective framed autonomy for hybrid business process representations that integrate both temporal and data-aware conditions. A key characteristic of framed autonomy is that the observed partial execution trace cannot be modified; instead, it must be replayed and accepted by the process frame exactly as recorded. We showed how such frames, composed of procedural (Petri nets) and declarative (MP-Declare) components, can be translated into 1-clock guarded finite-state automata and subsequently reduced to a numeric temporal planning problem. We implemented the proposed technique in a prototype system and evaluated its scalability over a collection of synthetic process frames combining real-time constraints, data conditions, and procedural structure. The presented approach may support automated identification of optimal termination points for ongoing process instances or facilitate sophisticated what-if analysis in which process analysts explore alternative partial executions as hypotheses.

The experimental results demonstrate how different sources of complexity, such as the expressiveness of declarative constraints, the degree of parallelism in the procedural model, and the length or nature of the prefix, affect grounding and search performance. For process frames of moderate size, our approach proves feasible, providing planning-based continuations that satisfy the multi-perspective constraints while minimizing repair costs. These results motivate further research on optimizing the grounding phase, which remains the most computationally intensive component of the approach, particularly as the structural complexity of process frames increases.

Several additional research directions emerge from this work. First, the development of an optimized grounding algorithm is essential for improving scalability and supporting increasingly complex hybrid process frames. Second, resource constraints could be incorporated into the framing mechanism, enabling *resource scheduling* scenarios in which a resource pool is treated as an explicit element of the process constraints. This would allow the planner to account for resource availability, contention, and allocation decisions during autonomy deliberation.

Another promising line of research concerns the integration of *Signal Temporal Logic* (STL), a formalism for specifying quantitative temporal properties over continuous, real-valued signals. External factors, such as weather conditions, machine sensor readings, patient vitals, or environmental measurements, may significantly influence process execution. Enriching the process frame with STL constraints would allow an ABPMS to react dynamically to such exogenous signals, enabling fully context-aware autonomy that adapts to changes in the operational environment.

While multi-perspective framed autonomy advances the state of the art by supporting hybrid process frames equipped with temporal and data conditions, it does not yet address the inherent variability and uncertainty characteristic of real-world executions. Many operational environments exhibit non-deterministic behavior that is naturally captured using *stochastic* or *probabilistic* process models. Extending our framework to incorporate probabilistic planning techniques would enable reasoning under uncertainty, allowing the system to recommend actions based not only on constraint satisfaction but also on expected utility.

Finally, recent advances in Large Language Models (LLMs) offer an opportunity to design a rich, conversational interface for the planning-based autonomy engine. LLMs can support natural-language explanations of planning decisions, facilitate interactive what-if analysis, and enable users to provide high-level goals or constraints through free-text queries. Such an interface would greatly enhance explainability, usability, and humanAI collaboration, key requirements for next-generation AI-augmented Business Process Management Systems.

## Declaration on Generative AI

The author(s) have not employed any Generative AI tools.

## Appendix A. Inputs of the Experimental Evaluation

The procedural models used in the experimental evaluation are given in Figs. A.16 to A.19. The model in Fig. A.16 (referred to as '0And' in the experimental evaluation) contains no parallelism. Each following model increases the level of parallelism by converting one XOR structure into an AND structure one at a time.

The declarative constraints used in the evaluation are the following:

1. Response[ActivityG, ActivityP] $|A.int > 10|T.int < 10|2, 5, h$
2. Precedence[ActivityF, ActivityD] $|A.cat$ is $c1|T.cat\ is\ c2|2, 5, h$
3. Not Response[ActivityM, ActivityO] $|A.int > 20|T.int < 10|2, 5, h$
4. Absence[ActivityN] $|A.int < 20|0, 100, h$
5. Chain Response[ActivityA, ActivityC] $|A.int < 5|T.cat$ is $c1|2, 5, h$
6. Existence[ActivityJ] $|A.cat$ is $c2|0, 100, h$
7. Response[ActivityH, ActivityQ] $|A.int > 10|T.int > 10|2, 5, h$

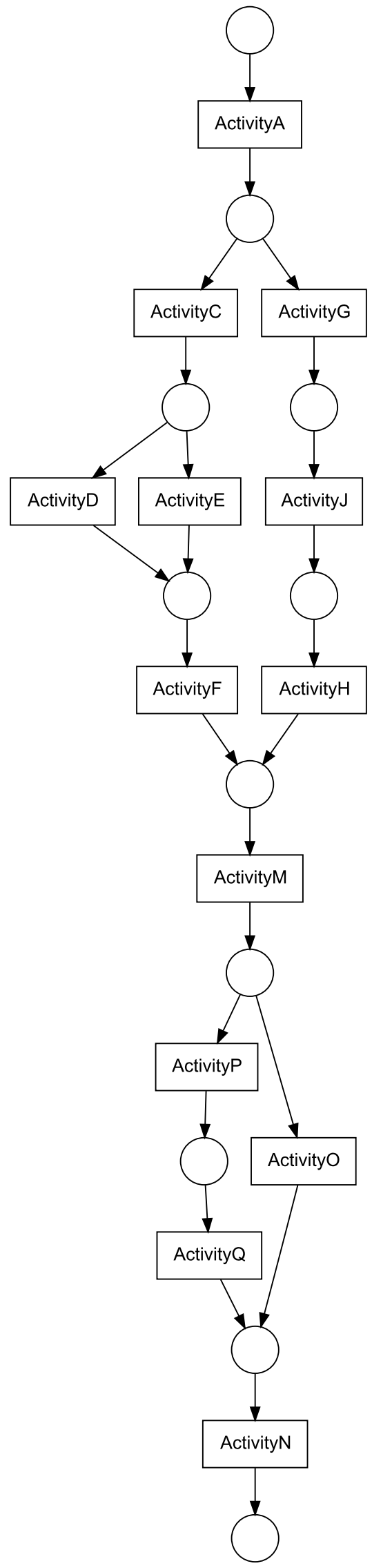

Figure A.16: Model 0And, containing three XOR structures and no AND structures.

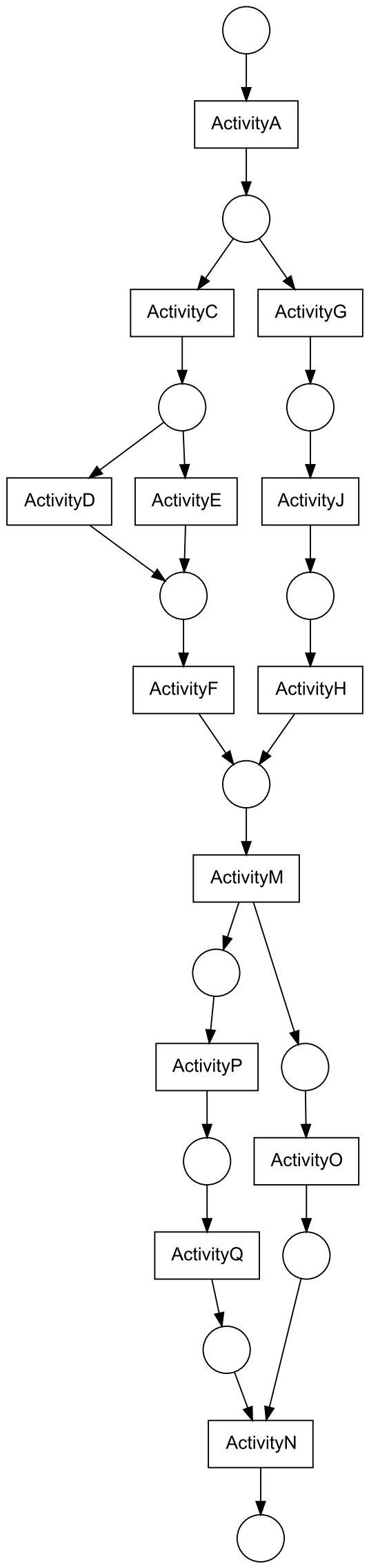

Figure A.17: Model 1And, containing two XOR structures and one AND structure.

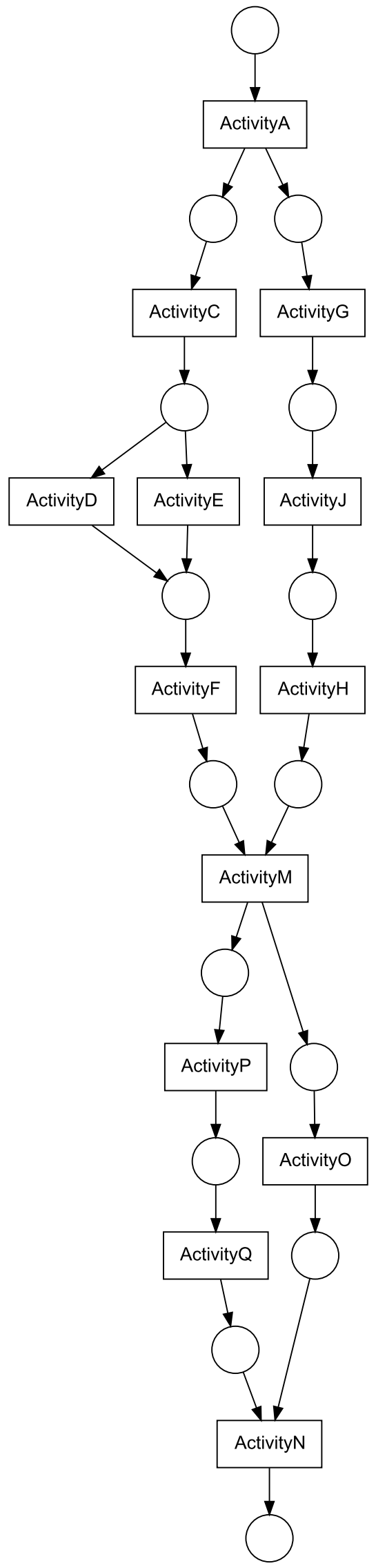

Figure A.18: Model 2And, containing one XOR structure and two AND structures.

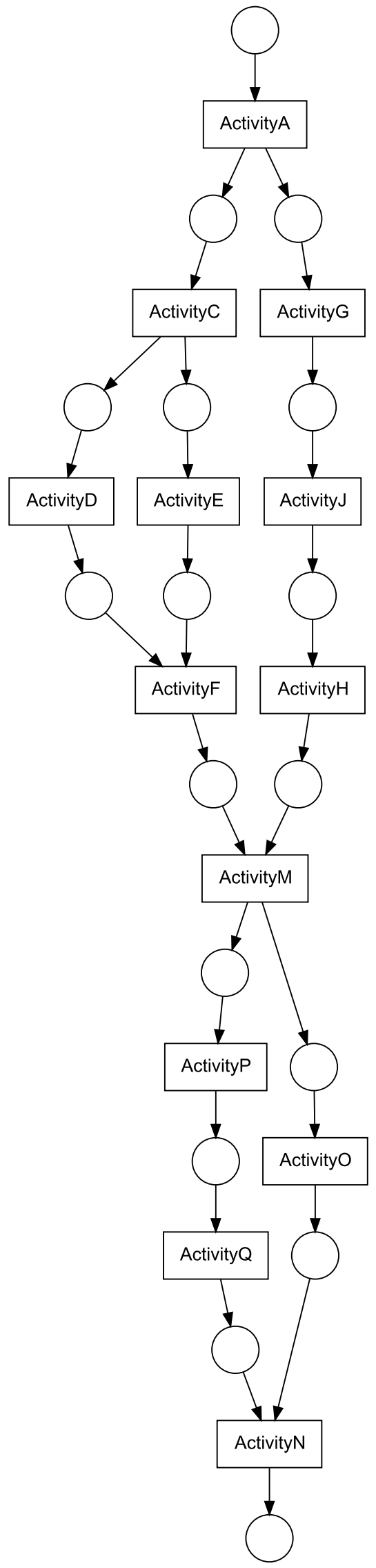

Figure A.19: Model 3And, containing no XOR structures and three AND structures.